%% file: acl_latex.tex
\pdfoutput=1

\documentclass[11pt]{article}

\usepackage[preprint]{acl}

\usepackage{times}
\usepackage{latexsym}
\usepackage[T1]{fontenc}
\usepackage[utf8]{inputenc}
\usepackage{microtype}
\usepackage{inconsolata}
\usepackage{graphicx}
\usepackage{listings}
\usepackage{amsfonts}

\usepackage{amsmath}
\usepackage{subcaption} 
\usepackage{xparse}
\usepackage{dblfloatfix}

\NewDocumentCommand{\codeword}{v}{%
\texttt{\textcolor{black}{#1}}%
}
\NewDocumentCommand{\codewordBlue}{v}{%
\texttt{\textcolor{blue}{#1}}
}

\title{Bridging Linguistic Structure and Mechanistic Interpretability for Conceptual Interpretation in Language Models}

\author{Nura Aljaafari$^{1\dagger}$,~ Danilo S. Carvalho$^{3}$,~ Andr\'{e} Freitas$^{1,2,3}$ \\
  $^{1}$ Department of Computer Science, University of Manchester, United Kingdom\\
  $^{2}$ Idiap Research Institute, Switzerland\\
  $^{3}$ CRUK National Biomarker Centre, University of Manchester, United Kingdom\\
  \texttt{\{firstname.lastname\}@[postgrad.]$^{\dagger}$manchester.ac.uk}}

\begin{document}
\maketitle

\begin{abstract}
Understanding how language models compose meaning from linguistic input remains a central problem in interpretability research. Mechanistic studies have attributed functional roles to core transformer components; however, these findings derive largely from factual retrieval settings. Whether the same mechanisms support \textit{conceptual interpretation}, the compositional mapping from definitional expressions to abstract meaning, remains insufficiently characterised. We introduce \textit{DSRA} (Definitional Semantic Role Analysis), a methodology that applies causal tracing within the reverse dictionary task and augments restoration traces with definitional semantic roles (DSRs) grounded in Argument Structure Theory. This linguistic overlay identifies which compositional functions (e.g., genus, differentia quality) are associated with high-recovery states, extending activation patching beyond token-level localisation. Applied to GPT-J-6B (English) and BERTIN GPT-J-6B (Spanish), the results show that MLP layers associate content-bearing tokens with high-specificity DSR categories in early layers, MHA layers distribute integration across middle-to-upper layers with concentration at the final token, and hidden states aggregate information in upper layers. Alignment between restored states and DSR categories indicates systematic correspondence between internal activations and definitional structure, with consistent localisation patterns across both languages.
\end{abstract}

\input{introduction}
\input{background}

\input{methodology}
\input{empirical}
\input{related_work}
\input{conclusion}
\input{acknowledgements}

\bibliography{acl_latex}

\appendix
\input{appendix}
\end{document}

%% file: introduction.tex
\section{Introduction}\label{sec:introduction}

Conceptual interpretation maps lexical items and syntactic structures to abstract meaning representations~\citep{brown2006encyclopedia,hirst1987semantic}. It supports sentence-level understanding and downstream tasks such as textual entailment and question answering. Transformer-based LMs achieve strong performance on these tasks, yet most evaluations focus on external accuracy and provide limited evidence about the internal computations that construct meaning. Consequently, the mechanisms by which models compose meaning from linguistic input remain insufficiently understood.

Mechanistic interpretability has identified stable functional roles for core transformer components. MLP layers have been characterised as key--value memory stores~\citep{geva-etal-2021-transformer,dai-etal-2022-knowledge}; attention heads within MHA modules often aggregate information toward the final-token position~\citep{Da2021AnalyzingCE}; and upper layers concentrate representations that are critical for prediction~\citep{li2022unified}. Sparse autoencoders further decompose superimposed activations into approximately monosemantic features~\citep{huben2023sparse,shu-etal-2025-survey}. However, these results are derived predominantly from factual retrieval prompts (e.g., ``Eiffel Tower is located in [Paris]''). Such settings establish component-level behaviours, but they do not determine whether, or how, these behaviours support compositional conceptual interpretation, in which meaning depends on integrating definitional structure, thematic roles, and predicate--argument relations.

\begin{figure*}[t]
  \includegraphics[width=\linewidth]{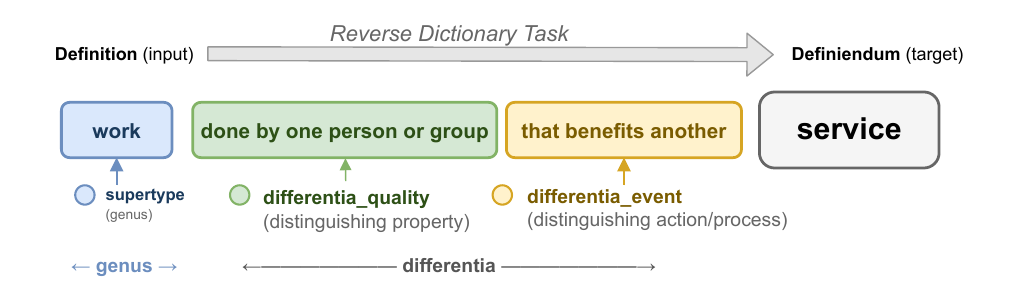}
  \caption{Definitional semantic labelling for the term \textit{service}. Each coloured span is annotated with its Definitional Semantic Role (DSR) following the taxonomy of~\citet{silva-etal-2016-categorization}. The DSR overlay provides an interpretive lens for assessing whether causally influential internal states align with definitional functions.}
  \label{fig:service_dsr}
\end{figure*}

We introduce DSRA (Definitional Semantic Role Analysis), a methodology for studying definition-based inference in the reverse dictionary task. DSRA applies causal tracing~\citep{meng2022locating} to localise internal states that are causally influential for predicting the \textit{definiendum}, and it interprets restoration traces using definitional semantic roles (DSRs)~\citep{silva-etal-2016-categorization}. In the reverse dictionary task, a model receives a natural-language definition and predicts the single term it defines. For example, given ``work done by one person or group that benefits another,'' a model should predict \textit{service} (Figure~\ref{fig:service_dsr}). Definitions are well suited to this analysis because they exhibit systematic compositional structure: a broader category (the \textit{genus}, e.g., \textit{work}) is refined by distinguishing properties (the \textit{differentia}, e.g., \textit{done by one person or group that benefits another}), providing a controlled setting for studying how models integrate structured meaning into a single prediction.

The DSR taxonomy decomposes each definition into labelled spans, including \textit{supertype} (the immediate broader category), \textit{differentia\_quality} (distinguishing intrinsic property), and \textit{differentia\_event} (distinguishing action or process), which correspond to the compositional functions of the constituents. By overlaying these labels onto causal traces, DSR annotation supports analyses that move beyond token identity and test whether restored internal states align with definitional functions.

Three research questions guide the study. \textbf{RQ1}: What consistent localisation patterns do causal traces exhibit during definition-based inference, and how do these patterns generalise across languages? \textbf{RQ2}: Is functional differentiation observable across LM components (MLP, MHA, hidden states) during definition-based inference? \textbf{RQ3}: To what extent does grouping restored states by DSR category and part of speech reveal structured alignment between internal activations and definitional structure? To clarify what would count as evidence of structure, we state the corresponding baselines. If recovery under causal tracing were not systematically localised, we would expect broadly uniform layer--token patterns and limited cross-lingual stability (RQ1). If component type were not functionally differentiated, we would expect MLP, MHA, and hidden-state restorations to exhibit similar localisation profiles (RQ2). If recovery were not aligned with definitional structure, grouping top-$K$ restored states by DSR (and PoS) would not show systematic representation of particular roles (RQ3).

The contributions are threefold. First, DSRA provides a controlled experimental framework linking causal tracing to formal lexical semantics through the reverse dictionary task. Second, the DSR overlay reveals regularities that are not apparent from token identity alone: \textit{differentia\_quality} accounts for over 31\% of the top-50 MLP restoration states, whereas \textit{supertype} and \textit{differentia\_event} each contribute approximately 6\%. Third, an integrated analysis across MLP, MHA, and hidden states in GPT-J-6B (English) and BERTIN GPT-J-6B (Spanish) shows consistent localisation patterns across both languages and clarifies how component-level behaviours jointly support definition-based inference.


%% file: background.tex
\section{Background}\label{sec:bg}

\textbf{Natural language definitions.} Natural language (NL) definitions provide a controlled setting for analysing conceptual interpretation in LMs. A definition consists of two parts: the \textit{definiendum} (the term being defined) and the \textit{definiens} (the expression that specifies its meaning). Dictionary-style definitions express the necessary and sufficient conditions and essential attributes of a term, and interpreting them requires syntactic processing, disambiguation, and semantic composition. At a high level, the goal of this work is to identify which parts of a definition contribute most to a model's prediction of the definiendum, and whether these contributions correspond to linguistically meaningful components such as category terms and distinguishing properties. Definitions provide a natural setting for this analysis because they encode structured meaning in a relatively controlled and interpretable form, enabling systematic comparison between internal model behaviour and linguistic structure.

The analysis focuses on \textit{intensional definitions}, which specify meaning through conditions governing term usage rather than by enumeration of instances. Intensional definitions commonly exhibit \textit{genus--differentia} structure~\citep{silva-etal-2016-categorization}: a broader category (the genus) is identified and then restricted by distinguishing properties (the differentia). For example, the term \textit{service}, defined as ``work done by one person or group that benefits another'', contains \textit{work} as genus and the remaining clause as differentia (Figure~\ref{fig:service_dsr}). This structure is systematic and linguistically principled, making it well-suited to controlled analysis of how models integrate structured meaning in internal representations.

\begin{figure*}[t]
  \includegraphics[width=\textwidth]{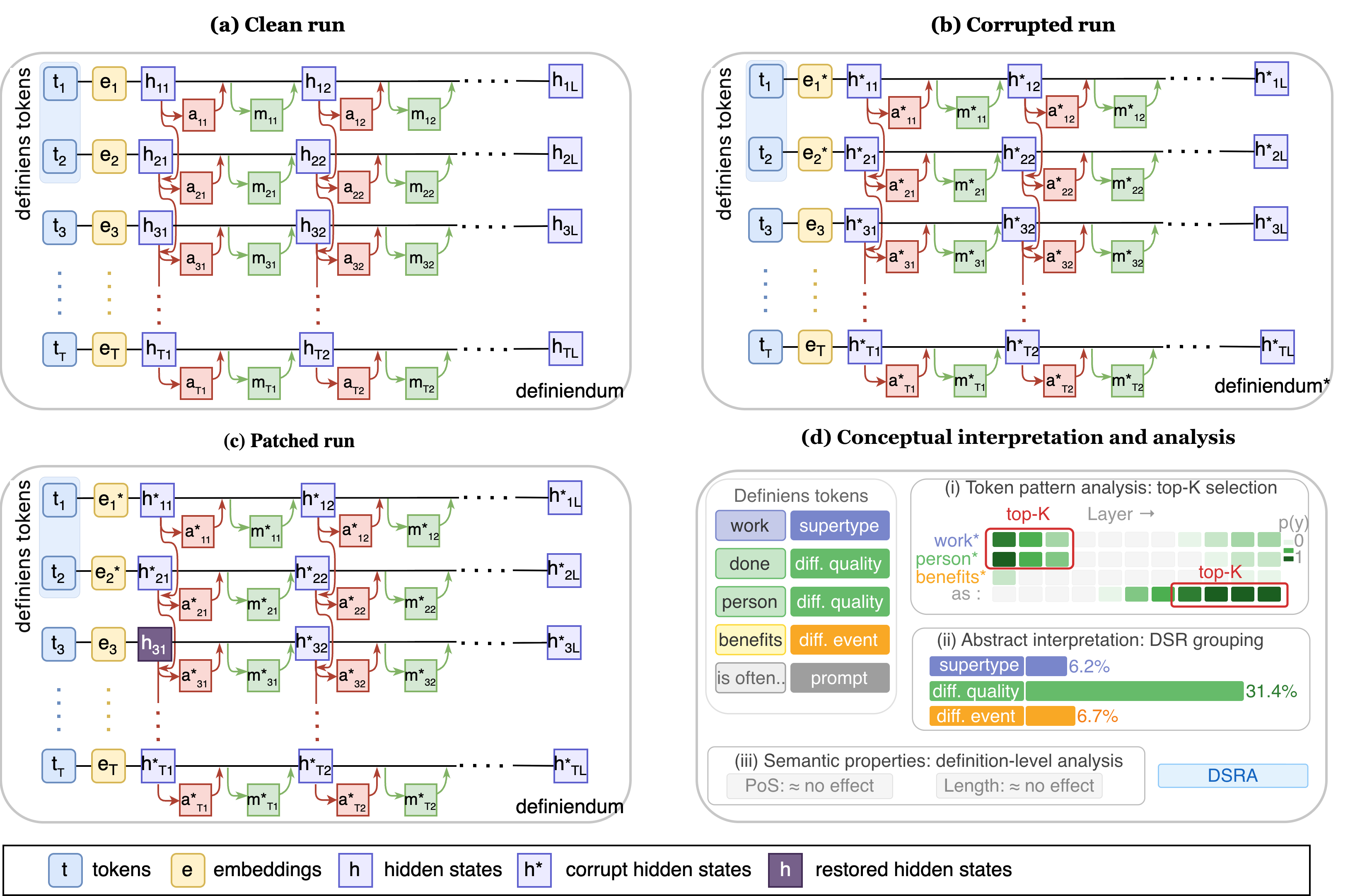}
  \caption{Overview of causal tracing and conceptual localisation in LMs. \textbf{(a)}~Clean run: the model processes the original definition. \textbf{(b)}~Corrupted run: Gaussian noise is added to definition-token embeddings. \textbf{(c)}~Patched run: selected states are restored from clean-run values. \textbf{(d)}~Conceptual interpretation analysis: restored states are grouped by token position, DSR label, and PoS. Model: GPT-J-6B (28 layers). Notation: $t$\,=\,tokens, $e$\,=\,embeddings, $h$\,=\,hidden states, $h^*$\,=\,corrupted states, $\bar{h}$\,=\,restored states, $a$\,=\,attention, $m$\,=\,MLP.}
  \label{fig:Causal_Trace_Approach}
\end{figure*}

\textbf{Conceptual composition.} To provide a formal basis for the DSR taxonomy used in this work, composition is examined under a Montagovian framework~\citep{partee1984compositionality}, in which the meaning of a complex expression is determined by the meanings of its parts and the way they are combined. Intuitively, a definition such as ``work done by one person or group that benefits another'' composes a category term (\textit{work}) with a set of restricting properties (\textit{done by one person\ldots}) to yield the meaning of the definiendum (\textit{service}). Sentence representation is formalised using Argument Structure Theory (AST)~\citep{jackendoff1992semantic,levin1993english, rappaport2008english}, in which a predicate $p$ is associated with arguments $arg_i$, each carrying a thematic role $r_i$ (see App.~\ref{appendix:alt_defs_generation} Table~\ref{tab:semantic_roles_def} for the full DSR taxonomy). Following \citet{silva-etal-2016-categorization}, a definiens statement $s_{\text{definiens}}$ is modelled as a composition of predicate--argument structures. In latent space:

\begin{equation}
    [[s_{\text{definiens}}]] = \underbrace{t_1({c_1}, {r_1})}_{\text{ARG0-genus}} \oplus \dots \oplus \underbrace{t_i({c_i}, {r_i})}_{\text{ARGN-differentia-quality}},
\end{equation}
where $[[s_{\text{definiendum}}]] = [[s_{\text{definiens}}]]$ and $t_i({c_i}, {r_i}) = c_i \otimes r_i$ follows compositional-distributional semantics notation~\citep{smolensky2006harmonic,clark2008compositional}. The operator $\otimes$ binds lexical content to thematic role, and $\oplus$ composes lexical-semantic units into definition-level meaning.

\textbf{Reverse dictionary task.} A valid interpretation of a definition assigns meanings to the elements of the definiens such that the resulting representation evaluates correctly for the definiendum. For generative LMs, this corresponds to next-token prediction with the definition as input and the definiendum as the expected continuation.\\

\textbf{Knowledge localisation.} Methods including ROME~\citep{meng2022locating}, MEMIT~\citep{meng2023memit}, and PMET~\citep{li2024pmet} modify factual associations by intervening on MLP representations. SAEs further reveal approximately monosemantic features~\citep{huben2023sparse,shu-etal-2025-survey}. These findings motivate investigation of whether similar tools can characterise compositional conceptual interpretation beyond factual retrieval.

%% file: methodology.tex
\section{Methodology}\label{sec:approach}

\subsection{Localisation via Causal Tracing}\label{sec:causal_tracing}
We adapt the causal tracing procedure of ROME~\citep{meng2022locating} to the reverse dictionary task in order to localise internal states that are causally implicated in predicting the \textit{definiendum}. Figure~\ref{fig:Causal_Trace_Approach} summarises the clean, corrupted, and patched executions and the subsequent analyses. Let $\mathcal{M}$ denote a transformer with $L$ layers that processes an input sequence $X$ comprising a natural-language definition embedded in a fixed prompt template (e.g., ``\textit{definition} is often referred to as:''). The target is the correct definiendum token $y$ predicted at the final position. For token position $t$ at layer $l$, we denote the residual-stream representation as $h^l_t$. For concision, we write the residual update as:
\begin{equation}
    h^l_t = h^{(l-1)}_t + \text{MHA}^l_t + \text{MLP}^l_t,
    \label{equation:hidden_states}
\end{equation}
where $\text{MHA}^l_t$ and $\text{MLP}^l_t$ denote the residual contributions of the attention and feed-forward sublayers at layer $l$ and position $t$, with all layer normalisation and architectural details inherited from the base model.

Causal tracing proceeds in three stages. In the \textit{clean execution}, the model processes $X$ and caches internal activations. In the \textit{corrupted execution}, Gaussian noise is injected into the embedding representations of the definition span, yielding a corrupted input $X^*$ that degrades the output distribution. In the \textit{patched execution}, selected internal states from the corrupted run are restored to their corresponding clean-run values. Recovery toward the clean prediction indicates that the restored states are causally sufficient for predicting $y$ under this intervention~\citep{meng2022locating}; that is, restoring these states is enough to recover the output. Conversely, limited recovery suggests that the restored states are not causally implicated. While this establishes causal sufficiency at the level of individual component restorations, it does not characterise the full causal graph of interactions between components. The span-delimitation rule, noise specification, and restoration operators are provided in Appendix~\ref{appendix:implementation_details}.

\subsection{Recovery Metric and Semantic Alignment}\label{sec:trace}
Let $p_y(X)$ denote the softmax probability assigned to the correct definiendum token $y$ under the clean execution, and let $p_y(X^*)$ denote the corresponding probability under the corrupted execution. For component $k \in \{\text{MLP}, \text{MHA}, \text{hidden}\}$, restoration at token position $i$ and layer index $j$ yields probability $p_y(X^{*,\text{patch}(i,j,k)})$. Following \citet{meng2022locating}, we define the component-specific recovery score:
\begin{equation}
\mathbf{T}^{(k)}_{ij} =
\frac{p_y(X^{*,\text{patch}(i,j,k)}) - p_y(X^*)}
     {p_y(X) - p_y(X^*)}.
\label{eq:recovery}
\end{equation}
This normalisation measures the fraction of corruption-induced degradation recovered by restoring a given internal state. Values are clipped to $[0,1]$ for numerical stability, consistent with prior work. Each input is evaluated across 10 independent corrupted executions, and recovery scores are averaged.

Restoration granularity follows \citet{meng2022locating}. For MLP and MHA components, restoration is applied over a contiguous window of 10 layers centred at layer index $j$; accordingly, $\mathbf{T}^{(\text{MLP})}_{ij}$ and $\mathbf{T}^{(\text{MHA})}_{ij}$ summarise the effect of restoring the corresponding component across that layer window at token position $i$. For hidden states, restoration targets individual layer--token pairs. A formal specification of the restoration operators and normalisation is provided in App.~\ref{appendix:formal_background}. Quantitative summaries use the top-$K$ restored states per example, defined as the $(i,j)$ pairs with the largest $\mathbf{T}^{(k)}_{ij}$ values. Unless stated otherwise, $K{=}50$ is used for reporting, and $K{=}10$ is reported for comparison. To aggregate traces across definitions of varying length, traces are length-normalised by resampling along the token dimension; details of the resampling grid and the length band are given in App.~\ref{appendix:length_norm}.
For semantic alignment, tokens in the English setting are assigned DSR labels~\citep{silva-etal-2016-categorization} and part-of-speech (PoS) tags. Top-$K$ restored states are grouped by these categories, relating recovery patterns $\mathbf{T}$ to span-level semantic functions in the definition. Spanish definitions are analysed using PoS tags only because DSR labels are unavailable for SWN.

\subsection{Datasets and Experimental Setup}\label{sec:setup}
The experimental design requires (i) a task with a verifiable target that elicits compositional semantic processing, (ii) datasets that provide structured definitional resources, and (iii) models that are sufficiently large to support per-example causal tracing. The reverse dictionary task satisfies (i) because it requires integrating multi-part definitional structure (e.g., genus, distinguishing properties, and relational content) into a single prediction and provides an unambiguous correctness criterion. This controlled formulation isolates compositional semantic processing from confounds such as multi-token generation variability, making it well-suited for per-example causal tracing at this scale. Whilst this task setting does not preclude the possibility that some high-recovery states reflect lexical co-occurrence rather than structured compositional processing, the use of DSR-based grouping enables analysis of whether recovery patterns align with functional definitional roles beyond individual token identity. Full dataset construction, filtering, and prompt selection are provided in Appendix~\ref{appendix:dataset_creation}. The extent to which these patterns generalise to other tasks and model scales is discussed in Section~\ref{limitations}.

\paragraph{Datasets.} Two datasets are constructed from WordNet~\citep{miller-1994-wordnet} and Spanish WordNet~\citep{Gonzalez-Agirre:Laparra:Rigau:2012}, referred to as EWN and SWN respectively. EWN contains 8\,348 samples (80/20 train--test split), and SWN contains 7\,815 samples (30/70 train--test split). Each sample in EWN is augmented with DSR labels~\citep{silva-etal-2016-categorization}.\footnote{DSR annotations were limited to EWN due to the unavailability of a comparable annotation resource for Spanish.} In both settings, definienda are restricted to single tokens to ensure that prediction correctness is unambiguous. 

Causal tracing is computed on the subset of test definitions that the fine-tuned models predict correctly under clean (unperturbed) conditions; filtering criteria and counts are provided in Appendix~\ref{appendix:dataset_creation}. These resources were selected for their structured and consistent definitional format, which enables controlled compositional analysis; their known limitations are discussed in Section~\ref{limitations}.

\paragraph{Models.} The models are GPT-J-6B~\citep{gpt-j} for English and BERTIN GPT-J-6B~\citep{BERTIN-GPT} for Spanish, fine-tuned via LoRA~\citep{hu2021lora} and QLoRA~\citep{dettmers2024qlora} respectively. Both models share the same base architecture, which enables direct comparison of localisation patterns across languages while controlling for architectural variation; full specifications are provided in App.~\ref{appendix:model_cards}.

\begin{figure*}[t]
    \centering
    \begin{subfigure}{0.33\textwidth}
        \centering
        \includegraphics[width=\linewidth]{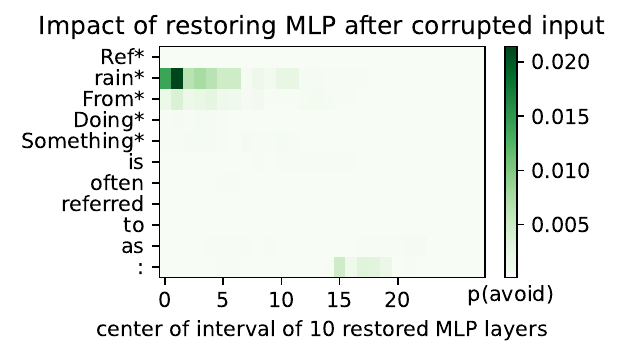}
        \caption{} \label{fig:restored_mlp_1}
    \end{subfigure}
    \hfill
    \begin{subfigure}{0.32\textwidth}
        \centering
        \includegraphics[width=\linewidth]{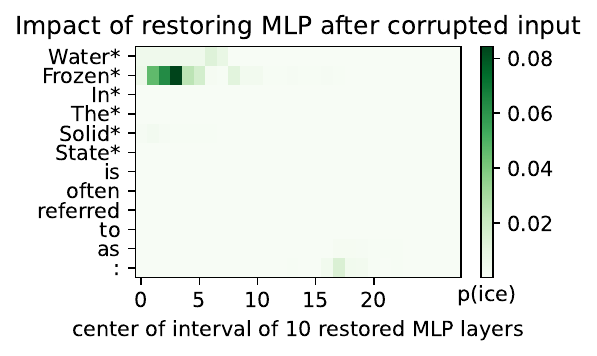}
        \caption{} \label{fig:restored_mlp_2}
    \end{subfigure}
    \hfill
    \begin{subfigure}{0.33\textwidth}
        \centering
        \includegraphics[width=\linewidth]{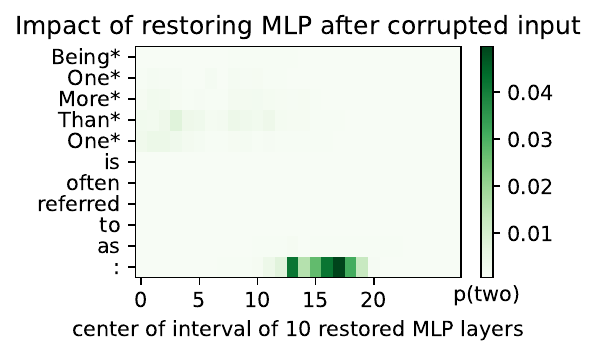}
        \caption{} \label{fig:restored_mlp_3}
    \end{subfigure}
    \caption{Causal traces for MLP restoration (GPT-J-6B, EWN). Each heatmap shows the normalised recovery score $\mathbf{T}^{(\text{MLP})}$ when restoring a sliding window of 10 MLP layers at each input token (y-axis; asterisks denote corrupted definition tokens). (a)~\textit{avoid}: strong early-layer recovery at content-bearing tokens. (b)~\textit{ice}: a similar early-layer pattern. (c)~\textit{two}: recovery shifts toward upper layers and the final token when definition tokens provide weaker lexical cues.}
    \label{fig:restored_mlp}
\end{figure*}

%% file: empirical.tex
\section{Empirical Analysis}\label{sec:empirical}
Figures~\ref{fig:restored_mlp}, \ref{fig:restored_attention}, and~\ref{fig:restored_hidden} follow the visualisation approach of \citet{meng2022locating}. The y-axis indexes input tokens (asterisks indicate corrupted definition tokens), and the x-axis indexes layer positions from $0$ to $L{-}1$. Each cell reports a normalised recovery score in $[0,1]$. Unless stated otherwise, quantitative summaries are based on the top-$K$ restored states per example for $K \in \{10,50\}$. To aggregate traces across definitions of varying length, global trends use length-normalised resampling over a common length band covering approximately 80\% of examples (Appendix~\ref{appendix:length_norm}). Layer-index distributions and additional grouped analyses are reported in Appendix~\ref{appendix:supplementary_results}.

\subsection{MLP: Content Association in Early and Upper Layers}\label{sec:mlp_summary}
\paragraph{Localisation pattern (RQ1).}
MLP traces show two recurring localisation regimes. High-recovery states appear in early layers at content-bearing tokens that are lexically or semantically related to the \textit{definiendum} (Figure~\ref{fig:restored_mlp_1} and~\ref{fig:restored_mlp_2}). When the definition provides weaker lexical cues, recovery shifts toward upper layers and concentrates at the final token position (Figure~\ref{fig:restored_mlp_3}). Aggregated layer-index distributions exhibit a bimodal pattern with concentration in layers 0--5 and 17--21 (Appendix~\ref{appendix:supplementary_results}). This structured concentration contrasts with a uniform distribution that would indicate limited localisation. The pattern is consistent with key--value retrieval behaviour reported for MLP layers~\citep{geva-etal-2021-transformer,dai-etal-2022-knowledge}. Evidence is also consistent with early tokenisation effects in which split tokens contribute to recovery when restored (Figure~\ref{fig:restored_mlp_1}; see also Appendix Figures~\ref{subfig:Alternative_definitions_MLP_1} and~\ref{subfig:Alternative_definitions_MLP_2})

\paragraph{Component differentiation (RQ2).}
The bimodal layer concentration for MLP differs from the unimodal concentration observed for MHA (Section~\ref{sec:mha_summary}) and hidden states (Section~\ref{hidde_states_analysis}), indicating differentiation across component types.

\paragraph{DSR alignment (RQ3).}
DSR-labelled MLP traces are provided in Appendix~\ref{appendix:dsr_traces}. Quantitatively, \textit{differentia\_quality} accounts for over 31\% of the top-50 restored states, followed by \textit{differentia\_event} and \textit{supertype} at approximately 6.7\% and 6.2\% respectively (Table~\ref{tab:dsr_distribution}). A null model in which DSR proportions match corpus-level span frequencies would predict substantially weaker concentration; the observed dominance of \textit{differentia\_quality} therefore supports systematic correspondence between high-impact states and definitional structure.

\begin{table}[t]
\centering
\small
\begin{tabular}{lrrr}
\hline
\textbf{DSR Label} & \textbf{MLP} & \textbf{MHA} & \textbf{Hidden} \\
\hline
\multicolumn{4}{c}{\textit{Top 50 states (\%)}} \\
\hline
diff.~quality & 31.4 & 34.0 & 32.0 \\
diff.~event   &  6.7 &  7.0 &  7.0 \\
supertype     &  6.2 & $<$1.0 & $<$1.0 \\
other DSR     &  6.7 &  4.0 & 10.0 \\
prompt        & 49.0 & 54.0 & 51.0 \\
\hline
\end{tabular}
\caption{Distribution of DSR labels among top-50 causal tracing states per component type (EWN, GPT-J-6B, $n{=}818$ correctly predicted test samples). For each example and component, top-50 indices are the $(i,j)$ pairs with the largest $\mathbf{T}^{(k)}_{ij}$ values; percentages are averaged across samples. Prompt tokens are included for completeness.}
\label{tab:dsr_distribution}
\end{table}

\subsection{MHA: Distributed Integration in Middle-to-Upper Layers}\label{sec:mha_summary}
\begin{figure*}[t]
    \centering
    \begin{subfigure}[t]{0.33\textwidth}
        \centering
        \includegraphics[width=\linewidth]{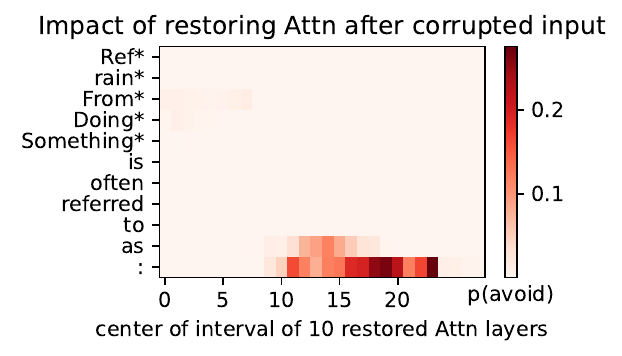}
        \caption{} \label{fig:restored_attn_1}
    \end{subfigure}
    \hfill
    \begin{subfigure}[t]{0.32\textwidth}
        \centering
        \includegraphics[width=\linewidth]{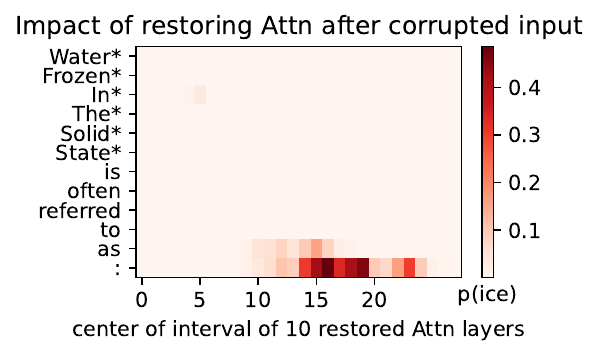}
        \caption{} \label{fig:restored_attn_2}
    \end{subfigure}
    \hfill
    \begin{subfigure}[t]{0.33\textwidth}
        \centering
        \includegraphics[width=\linewidth]{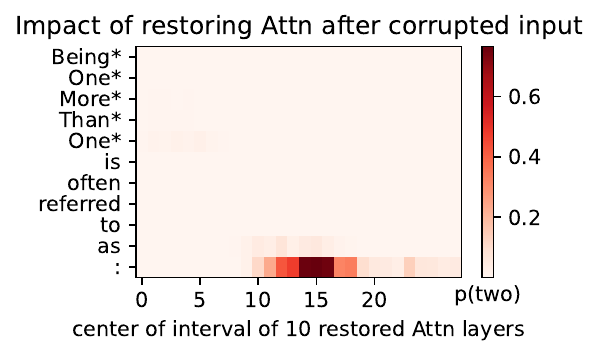}
        \caption{} \label{fig:restored_attn_3}
    \end{subfigure}
    \caption{Causal traces for MHA restoration (GPT-J-6B, EWN). Each heatmap shows the normalised recovery score $\mathbf{T}^{(\text{MHA})}$ when restoring a sliding window of 10 MHA layers (x-axis: window centre, layers 0--27) at each input token (y-axis; asterisks denote corrupted definition tokens). Influential states concentrate at the final token across middle-to-upper layers.}
    \label{fig:restored_attention}
\end{figure*}

\paragraph{Localisation pattern (RQ1).}
MHA traces consistently concentrate at the final token, with influential states distributed across middle-to-upper layers (12--25) (Figure~\ref{fig:restored_attention}). Aggregated layer-index distributions show unimodal, negatively skewed concentration (Appendix~\ref{appendix:supplementary_results}), contrasting with the bimodal MLP pattern.

\paragraph{Component differentiation (RQ2).}
Prompt tokens dominate among top-10 states, whereas content-bearing roles enter the high-impact set when considering top-50 states (Table~\ref{tab:dsr_distribution}). This pattern is consistent with gradual information aggregation during definition-based inference, with definition content contributing more strongly at broader restoration thresholds.

\paragraph{DSR alignment (RQ3).}
DSR-labelled MHA traces are provided in Appendix~\ref{appendix:dsr_traces}. The shift from prompt-dominated top-10 states to DSR-represented top-50 states indicates progressive integration of definition content.

\subsection{Hidden States: Upper-Layer Concentration}\label{hidde_states_analysis}
\begin{figure*}[t]
  \centering
  \begin{subfigure}{0.33\textwidth}
        \centering
         \includegraphics[width=\linewidth]{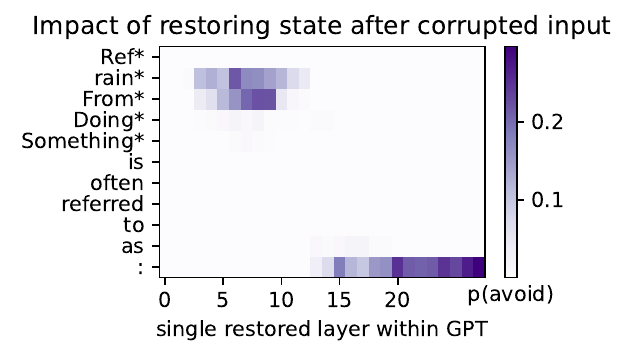}
        \caption{} \label{fig:restored_hidden_1}
  \end{subfigure}\hfill
  \begin{subfigure}{0.33\textwidth}
    \centering
    \includegraphics[width=\linewidth]{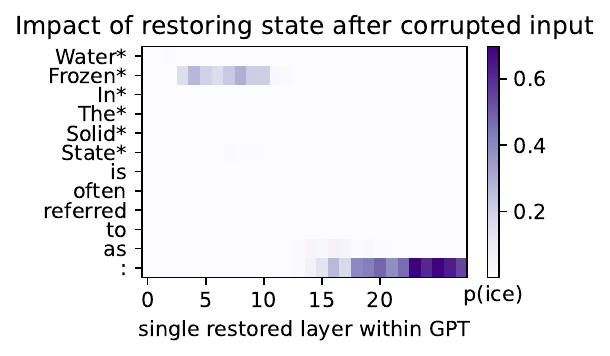}
        \caption{} \label{fig:restored_hidden_2}
  \end{subfigure}\hfill
  \begin{subfigure}{0.33\textwidth}
        \centering
        \includegraphics[width=\linewidth]{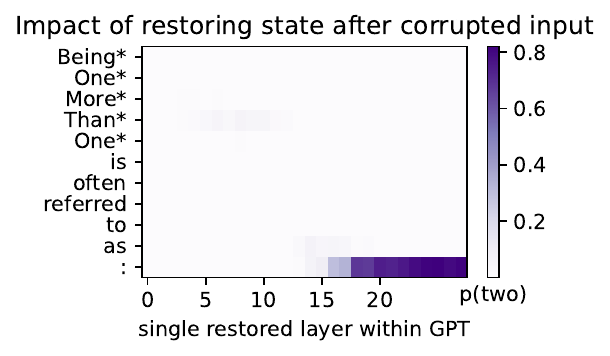}
        \caption{} \label{fig:restored_hidden_3}
   \end{subfigure}
  \caption{Causal traces for hidden-state restoration (GPT-J-6B, EWN). Each heatmap shows the normalised recovery score $\mathbf{T}^{(\text{hidden})}$ (colour scale, clipped to $[0,1]$) when restoring a single hidden state (x-axis: layer index 0--27) at each input token (y-axis; asterisks denote corrupted definition tokens). Influential states concentrate in upper layers and at the final token.}
  \label{fig:restored_hidden}
\end{figure*}
\begin{figure*}[t]
    \centering
    \includegraphics[width=\linewidth]{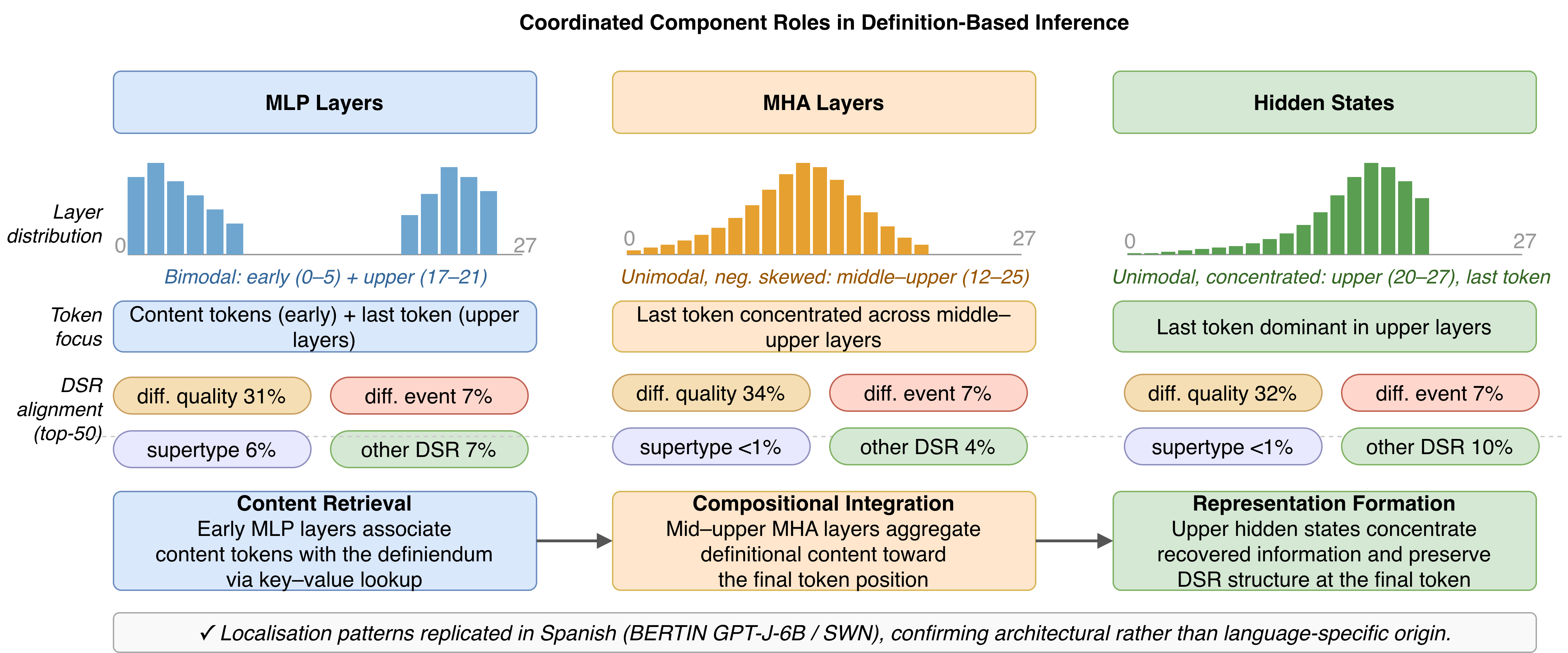}
    \caption{Coarse-grained synthesis of localisation results across MLP, MHA, and hidden-state components (GPT-J-6B, EWN). Bars schematically summarise layer-wise concentration patterns, and DSR percentages are taken from the top-50 analysis (Table~\ref{tab:dsr_distribution}). Arrows indicate a proposed directional account of information aggregation consistent with the observed traces.}
    \label{fig:synthesis}
\end{figure*}
\paragraph{Localisation pattern (RQ1).}
Hidden-state traces concentrate in upper layers (20--27) and at the final token (Figure~\ref{fig:restored_hidden}). Aggregated layer-index distributions show unimodal concentration in the top layers (Appendix~\ref{appendix:supplementary_results}). Tokens outside the prompt generally show low impact unless they are lexically related to the \textit{definiendum}.

\paragraph{Component differentiation (RQ2).}
The layer concentration for hidden states is more strongly upper-layer than the concentration for MHA, consistent with hierarchical aggregation into the final representation.

\paragraph{DSR alignment (RQ3).}
\textit{differentia\_quality} (32\%) and \textit{differentia\_event} (7\%) remain present among top-50 restored states (Table~\ref{tab:dsr_distribution}), indicating persistence of definitional information into the final representation. DSR-labelled MLP, MHA and hidden-state traces are provided in Appendix~\ref{appendix:dsr_traces}.

\subsection{Validation and Cross-Lingual Stability}\label{sec:validation}
\paragraph{Alternative definitions.}
Alternative definitions are generated as described in Appendix~\ref{appendix:alt_defs_generation}, with five alternatives per sample. Traces computed on correctly predicted alternatives preserve the broad localisation patterns while shifting specific high-impact positions across paraphrases (see Figures~\ref{fig:Alternative_definitions_MLP}--\ref{fig:Alternative_definitions_hidden} in Appendix~\ref{appendix:supplementary_results}).

\paragraph{Cross-lingual stability (RQ1).}
Experiments replicated on SWN with BERTIN GPT-J-6B yield localisation patterns consistent with the English analysis (see Figure~\ref{fig:spanish_combined_traces} in Appendix~\ref{appendix:supplementary_results}). Differences in the distribution of high-impact MLP positions are consistent with tokenisation variation while preserving component-level behaviour. Cross-lingual comparison is therefore limited to localisation patterns, as DSR-based semantic alignment is evaluated only in the English setting.

\subsection{Synthesis: A Coarse-Grained Account}\label{sec:outline}
Figure~\ref{fig:synthesis} summarises a coarse-grained account of coordinated behaviour across MLP, MHA, and hidden states. Overall, the results are inconsistent with simple baselines. MLP recovery concentrates in two regimes rather than exhibiting approximately uniform layer occupancy. In addition, localisation profiles differ across components: MLP bimodality contrasts with unimodal MHA and hidden-state concentration, indicating component differentiation. Under DSR grouping, \textit{differentia\_quality} is highly represented among high-impact states relative to frequency-matched expectations, suggesting a systematic correspondence between internal states and definitional structure rather than span prevalence alone. This correspondence is indicative of structured semantic alignment, though it does not by itself establish a complete account of compositional computation. These trends persist under paraphrase and replicate in Spanish, arguing against artefacts of surface form or language. MLP traces emphasise content association in early layers and shift toward upper-layer concentration at the final token when lexical cues are weak. MHA traces implement distributed aggregation centred on the final token across middle-to-upper layers. Hidden-state traces concentrate recovery in upper layers while retaining definitional structure under DSR grouping. Testing this account with finer-grained causal interventions is left to future work.




%% file: related_work.tex
\section{Related Work}\label{sec:related}
\textbf{Mechanistic interpretability (MI).} MI decomposes neural networks into human-understandable computational structures by identifying functional roles for internal components~\citep{conmy2023towards}. Probing studies have shown that transformer layers encode linguistic knowledge in a hierarchical progression~\citep{tenney2019bert}, establishing that internal representations align with structured linguistic abstractions. SAEs decompose superimposed activations into approximately monosemantic features~\citep{huben2023sparse,shu-etal-2025-survey} and have been applied to code correctness~\citep{tahimic2025mechanistic}, hallucination detection~\citep{xiong2026toward}, and causal representation learning~\citep{song2025llm}. While probing identifies the presence of linguistic information and SAEs isolate interpretable features, neither approach directly characterises how such information is composed and integrated across components during meaning construction. The present work addresses this gap by applying intervention-based analysis to compositional definitional processing.\\
\textbf{Activation patching and causal tracing.} Causal mediation analysis was first applied to transformer internals by \citet{vig2020investigating}, who treated attention heads as causal mediators of gender bias. Building on this framework, activation patching modifies and restores internal activations to identify causally relevant states~\citep{heimersheim2024use}. ROME~\citep{meng2022locating} and MEMIT~\citep{meng2023memit} apply this approach to knowledge localisation in MLP layers; Causal Scrubbing~\citep{chan2022causal} and Attribution Patching~\citep{nanda2023attribution} refine causal analysis at scale. Causal abstraction methods further align neural computations with formal semantic structures~\citep{geiger2021causal}, but have not been applied to definitional interpretation or lexical-semantic role taxonomies. Among available localisation methods, ROME-style causal tracing is adopted because it enables intervention-based localisation of internal representations at the component level, providing a direct and well-established approach for analysing decoder-only transformers at this scale~\citep{meng2022locating}, in contrast to gradient-based attribution methods, which estimate rather than directly intervene on internal states. DSRA extends this framework to conceptual interpretation by integrating DSR labels into restoration traces, enabling attribution at the level of semantic roles rather than token positions alone.\\
\textbf{Logit attribution and circuit analysis.} Logit attribution links predictions to input tokens via logit contributions~\citep{elhage2021mathematical} and has been applied to feed-forward layers~\citep{geva-etal-2022-transformer} and MHA heads~\citep{ferrando-etal-2023-explaining}. More recent attribution-based approaches, such as information flow routes~\citep{ferrando-voita-2024-information}, scale circuit discovery via single forward-pass attribution but, like logit attribution, rely on gradient-based estimates rather than direct causal intervention. Circuit analysis identifies functional subgraphs for specific behaviours~\citep{olsson2022context,wang2022interpretability,conmy2023towards}, typically focusing on individual attention heads and edge-level structures rather than the joint contribution of multiple component types. DSRA differs in examining MLP, MHA, and hidden-state components jointly under direct intervention, and relates recovery patterns to span-level definitional semantic structure rather than token-level logit contributions.
\textbf{Knowledge localisation in MLP layers.} MLP layers have been characterised as key--value memory stores~\citep{geva-etal-2021-transformer}, and \citet{dai-etal-2022-knowledge} demonstrated that specific MLP neurons directly affect factual recall, providing causal evidence for content-specific storage. These findings raise the question of whether MLP layers play analogous content-association roles during compositional conceptual interpretation, where meaning is derived from structured definitional input rather than a single factual association.\\
\textbf{Definition modelling and reverse dictionary.} Definition modelling, introduced by \citet{noraset2017definition}, established the task of generating dictionary definitions from word embeddings. \citet{bevilacqua2020generationary} extended this to pretrained transformers, demonstrating that fine-tuned models capture contextually appropriate definitional knowledge. More recently, \citet{xu-etal-2024-tip} used the reverse dictionary task as a probe for LLM conceptual inference, showing that model representations encode object categories and fine-grained features; however, this work relies on probing and in-context learning rather than causal intervention, and does not examine internal component-level behaviour. DSRs~\citep{silva-etal-2016-categorization} offer a structured framework for analysing definitions through genus--differentia structure and predicate--argument relations. This work addresses the gap at the intersection of these lines: to our knowledge, intervention-based causal analysis has not yet been applied to the reverse dictionary task to explore how definitional semantic structure may be reflected in transformer component behaviour. DSRA fills this gap by integrating the DSR lexical-semantic framework with mechanistic interpretability, moving beyond probing to causal localisation of definitional structure within transformer components.

%% file: conclusion.tex
\section{Conclusion}\label{sec:conclusion}
This work demonstrates the value of integrating formal linguistic structure into mechanistic interpretability through DSRA, which augments causal tracing with definitional semantic roles within the reverse dictionary task. The focus of the present work is localisation and semantic alignment; causal intervention for model editing is left to future work.

With respect to \textbf{RQ1}, causal traces exhibit consistent localisation patterns: MLP states cluster bimodally in early and upper layers, MHA states concentrate in middle-to-upper layers with a negatively skewed distribution, and hidden states concentrate in upper layers at the final token. Similar localisation patterns are observed in both English and Spanish. With respect to \textbf{RQ2}, the three component types exhibit distinct distributional signatures, bimodal (MLP), unimodal upper-skewed (MHA), and concentrated upper-layer (hidden), supporting functional differentiation in definition-based inference. With respect to \textbf{RQ3}, DSR grouping reveals that content-bearing semantic roles, particularly \textit{differentia\_quality}, account for a substantial fraction of high-impact states, suggesting that semantic role structure provides additional explanatory structure beyond token identity.

The principal methodological contribution is the DSR overlay for causal tracing, which functions as an interpretive lens bridging activation patching and formal lexical semantics. Future work may pursue finer-grained tracing at the level of individual attention heads, integrate SAEs for feature-level DSR alignment, and extend the framework beyond definitional interpretation to assess generality.

\section*{Limitations}\label{limitations}
This study is limited to a single scale of autoregressive transformer model and focuses on sentence-level meaning with single-token prediction, which does not fully reflect the complexities of conceptual processing across other tasks. The analysis examines each component type individually rather than investigating causal relationships between components, and MHA is studied as a whole rather than at the level of individual heads. DSR annotations are available only for the English dataset. Selecting causal tracing as the primary method may have precluded other valuable analytical perspectives. The experimental design relies on WordNet and Spanish WordNet, which provide a controlled setting but carry known limitations, such as sense distinctions that could be overly fine-grained, and hierarchical structures that do not always correspond to cross-linguistically stable semantic categories. These factors may affect DSR annotation quality and the generalisability of the observed localisation patterns. As WordNet definitions are more structured and templated than naturally occurring text, it remains an open question whether the observed alignment between internal activations and definitional structure would persist with explanatory dictionaries or corpus-derived definitions exhibiting freer linguistic variation.

\section*{Ethical Considerations}\label{ethics}
This study applies DSRA using WordNet, abstract labelling, and causal tracing to analyse conceptual interpretation mechanisms in transformer-based models. The use of WordNet carries potential risks, including biases inherent in the dataset. Abstract labelling may introduce distorted representations. The identified mechanisms could potentially be employed to alter model behaviour in unintended ways. Code and datasets are released to enable replication and verification.

%% file: acknowledgements.tex
\section*{Acknowledgements}
This work was partially funded by the SNSF project RATIONAL (200021E\_229196), the CRUK National Biomarker Centre, and supported by the Manchester Experimental Cancer Medicine Centre and the NIHR Manchester Biomedical Research Centre.

%% file: appendix.tex
\section{Dataset Creation, Filtering, and Prompt Selection}\label{appendix:dataset_creation}

Three eligibility conditions governed dataset and model choices. First, the task must elicit compositional semantic processing with a verifiable target. Second, the datasets must support structured definitional analysis, including span-level labels when available. Third, the models must be compatible with per-example causal tracing under the available compute budget. This section documents dataset construction, filtering, prompt selection, and trace aggregation.

\subsection{English WordNet (EWN)}\label{appendix:ewn_creation}
The EWN dataset was constructed from Princeton WordNet~\citep{miller-1994-wordnet} by selecting definienda consisting only of alphabetical characters and restricting definitions to a maximum of 25 words for computational efficiency. Samples were retained only if the definiendum matched a single token in the model vocabulary, confining predictions to single-token inferences. This yielded 8{,}348 examples, split 80/20 for training and testing.

\subsection{Spanish WordNet (SWN)}\label{appendix:swn_creation}
A similar procedure was followed for SWN using the Open Multilingual Wordnet~\citep{Gonzalez-Agirre:Laparra:Rigau:2012}. Alphabetic definienda were selected and their definitions were queried in Spanish WordNet. When a Spanish definition was available, it was selected, cleaned, and used. When no Spanish definition was found, the Spanish definiendum was translated into English, and an English WordNet entry was retrieved with a matching part of speech. The retrieved English definition was then translated into Spanish using the Googletrans library,\footnote{\url{https://github.com/ssut/py-googletrans}} and back translation was used as a validation check for translation correctness. This procedure yielded 7{,}815 examples. Definitions were restricted to a maximum of 25 words, covering 99\% of samples. The 30/70 train--test split was chosen to maximise the number of held-out definitions available for causal tracing while retaining sufficient training data for stable task performance under parameter-efficient fine-tuning.

\subsection{Preprocessing and Prompt Selection}\label{appendix:prompt_selection}
Inputs consist of the definition text with a prompt template appended. Three prompt formats were evaluated: a direct instruction template (\codeword{``Identify the term defined as: \{definition\}.''}), a question template (\codeword{``What word is described by this definition: \{definition\}?''}), and a completion-style template (\codeword{``\{definition\} is often referred to as:''}). Additional metadata (e.g., the part of speech of the definiendum) was also tested. Simple prompts without additional metadata yielded the highest accuracy. For causal tracing analysis, only examples correctly predicted by the fine-tuned models under clean (unperturbed) conditions are included. Additional preprocessing details are provided in the released code repository.\footnote{\url{anonymised}}

\subsection{Length Normalisation}\label{appendix:length_norm}
To aggregate traces across examples of varying definition length, the token dimension of each trace tensor is resampled to a fixed grid within a common length band covering approximately 80\% of examples. The band is defined by the empirical 10th and 90th percentiles of tokenised definition length under the analysis prompt. Traces outside the band are excluded from length-normalised aggregations. Resampling uses linear interpolation over token index to preserve coarse positional structure while enabling global layer-wise summaries.

\subsection{Definitional Semantic Role Annotation}\label{appendix:dsr}
DSR labels follow the taxonomy of \citet{silva-etal-2016-categorization}. Annotations are available for English WordNet definitions only; Spanish WordNet definitions are analysed without DSR alignment. Table~\ref{tab:semantic_roles_def} describes each DSR used in this work along with its definition.

\subsection{Alternative Definitions}\label{appendix:alt_defs_generation}
Five alternative definitions per sample are generated using GPT-3.5~\citep{openai2023chatgpt} with the following prompt instruction:
\textit{Given the word ``\{definiendum\}'' and its dictionary definition ``\{definition\}'', generate five alternative definitions that: (1) preserve the core meaning of the original definition, (2) do not include the word itself or direct morphological variants and (3) remain plausible as dictionary-style entries. Return only the five definitions, numbered 1--5.}

\noindent Alternatives are manually filtered to ensure semantic similarity to the original definition, plausibility as standalone dictionary entries, and absence of the definiendum in the generated text. Only alternatives predicted correctly by the fine-tuned models are included in causal tracing to match the clean-evaluation filter described in Appendix~\ref{appendix:prompt_selection}.

\begin{table}[!ht]
  \centering
  \small
  \begin{tabular}{lp{0.5\columnwidth}}
    \hline
    \textbf{Role} & \textbf{Definition} \\
    \hline
    supertype & The superclass of the immediate entity or an ancestor of it \\
    differentia quality & A fundamental, intrinsic attribute that distinguishes the entity from others within the same supertype \\
    differentia event & An action, state, or process in which the entity engages, necessary for differentiation from others within the same supertype \\
    event time  & The time at which a differentia event occurs \\
    event location  & The specific location of a differentia event \\
    quality modifier & A modifier indicating degree, frequency, or manner that refines a differentia quality \\
    origin location & The place of origin of the entity \\
    purpose & The primary objective behind the existence or occurrence of the entity \\
    associated fact & A fact connected to the existence or occurrence of the entity, serving as an incidental attribute \\
    accessory determiner & A determiner expression that does not restrict the scope of supertype--differentia \\
    accessory quality & A non-essential attribute for characterising the entity \\
    $[\text{role}]$ particle & A particle not contiguous with other role components \\
    \hline
  \end{tabular}
  \caption{Definitional semantic roles and their definitions following the taxonomy of~\citet{silva-etal-2016-categorization}.}
  \label{tab:semantic_roles_def}
\end{table}

\section{Implementation Details}\label{appendix:implementation_details}

\subsection{Causal Tracing Specification}\label{appendix:tracing_spec}
This subsection specifies the corruption and restoration operators used in causal tracing. The definition span is delimited as the token subsequence corresponding to \codeword{\{definition\}} in the prompt template (Appendix~\ref{appendix:prompt_selection}). Corruption injects independent Gaussian noise into the token embedding vectors of the delimited span in the corrupted execution. The noise scale and any exclusions (e.g., prompt tokens and the final prediction position) are fixed across experiments and reported with the released code.

For restoration, we follow \citet{meng2022locating}. For component $k \in \{\text{MLP}, \text{MHA}\}$, restoration at $(i,j,k)$ replaces the corresponding component activations at token position $i$ across a contiguous 10-layer window centred at $j$ with their clean-run values. For hidden-state restoration, patching replaces the residual-stream representation at a single layer-token pair. Boundary handling for the layer window and the exact operator definitions are provided in App.~\ref{appendix:formal_background}.


\subsection{Models and Selection Rationale}\label{appendix:model_cards}
Table~\ref{tab:model_hyperparameters} summarises the architecture hyperparameters of the models used. Model selection is constrained by the requirements of intervention-based causal tracing. First, the model must permit extraction and restoration of internal activations at the level of MLP, MHA, and residual-stream states, since the tracing protocol quantifies the causal contribution of restored component activations. Second, the model must operate at a scale for which component-level localisation patterns under causal interventions have been shown to be stable and interpretable in prior mechanistic work~\citep{geva-etal-2021-transformer,meng2022locating}. Third, to support cross-lingual comparison, the models must share the same base architecture so that observed differences in localisation patterns are not attributable to architectural variation.

GPT-J-6B is used for English because it is an open-weight, decoder-only transformer that satisfies these constraints while remaining computationally feasible for per-example causal tracing. GPT-J-6B is trained using Mesh Transformer JAX~\citep{mesh-transformer-jax} on the Pile dataset~\citep{pile}. BERTIN GPT-J-6B is used for Spanish because it instantiates the same GPT-J architecture and is trained on Spanish data (mC4-es-sampled (Gaussian))~\citep{BERTIN}. This pairing provides architectural control over layer depth, hidden dimensionality, and attention configuration (Table~\ref{tab:model_hyperparameters}), enabling cross-lingual analyses in which language and tokenisation constitute the primary sources of variation.

Models are loaded via the HuggingFace Transformers library in evaluation mode with gradients disabled; half-precision is used when required to satisfy memory constraints. Both models are fine-tuned for the reverse dictionary task using parameter-efficient adaptation via LoRA~\citep{hu2021lora}, with settings reported in Table~\ref{tab:lora_hyperparameters}. This selection therefore prioritises methodological validity for causal tracing, architectural comparability for cross-lingual analysis, and practical feasibility for exhaustive tracing at the level of individual examples.

\begin{table}[ht!]
  \centering
  \small
  \begin{tabular}{lp{0.5\columnwidth}}
    \hline
    \textbf{Hyperparameter} & \textbf{Value} \\
    \hline
    $n_\text{parameters}$ & 6{,}053{,}381{,}344 \\
    $n_\text{layers}$ & 28\textsuperscript{*} \\
    $d_\text{model}$ & 4096 \\
    $d_\text{ff}$ & 16384 \\
    $n_\text{heads}$ & 16 \\
    $d_\text{head}$ & 256 \\
    $n_\text{ctx}$ & 2048 \\
    $n_\text{vocab}$ & 50257 / 50400\textsuperscript{\textdagger} \\
    Positional Encoding & RoPE \\
    RoPE Dimensions & 64 \\
    \hline
  \end{tabular}
  \caption{Architecture hyperparameters for GPT-J-6B and BERTIN-GPT-J-6B. Both models share the same base architecture. \textsuperscript{*}Includes one embedding layer and 27 transformer blocks. \textsuperscript{\textdagger}50{,}257 used tokens / 50{,}400 vocabulary size including padding.}
  \label{tab:model_hyperparameters}
\end{table}

\begin{table}[ht!]
  \centering
  \small
  \begin{tabular}{lcc}
    \hline
    \textbf{Setting} & \textbf{GPT-J-6B} & \textbf{BERTIN-GPT-J-6B} \\
    \hline
    Epochs & 10 & 20 \\
    Batch size & 16 & 16 \\
    LoRA rank ($r$) & 64 & 16 \\
    LoRA $\alpha$ & 32 & 32 \\
    Dropout & 0.1 & 0.3 \\
    \hline
  \end{tabular}
  \caption{LoRA fine-tuning hyperparameters for the reverse dictionary task.}
  \label{tab:lora_hyperparameters}
\end{table}

\subsection{Experimental Setup}\label{appendix:setup}
Experiments were conducted on an NVIDIA RTX A6000 GPU using Python 3.11.7, PyTorch 2.2.2, Hugging Face Transformers 4.39.0, and PEFT 0.8.2. Data processing relied on Datasets 2.16.1, NumPy 1.26.3, Pandas 2.2.0, and scikit-learn 1.4.0. Text processing used NLTK 3.8.1 and spaCy 3.7.2.

\section{Formal Background and Tracing Protocol}\label{appendix:formal_background}

\subsection{Transformer Formalisation}\label{appendix:transformer}
We summarise the standard transformer architecture~\citep{vaswani2017attention} to establish notation used throughout the paper. A decoder-only transformer with $L$ layers processes an input sequence $X = (x_1, \ldots, x_T)$ by computing a sequence of hidden representations. At each layer $l \in \{1, \ldots, L\}$ and token position $t$, the hidden state is updated via the residual stream:
\begin{equation}
    h_t^{(l)} = h_t^{(l-1)} + \text{MHA}^l_t + \text{MLP}^l_t,
\end{equation}
where $h_t^{(0)}$ is the sum of the token embedding and positional encoding for $x_t$. The multi-head attention (MHA) term concatenates the outputs of $H$ independent attention heads:
\begin{equation}
    \text{MHA}^{l}_t = W^l_O \; \text{Concat}_{h=1}^{H}\!\left(\text{Attn}^{l,h}(h_{\le t}^{(l-1)})\right),
    \label{equation:mha}
\end{equation}
where each head computes scaled dot-product attention over all positions $\le t$ (causal masking) and $W^l_O \in \mathbb{R}^{d_{\text{model}} \times d_{\text{model}}}$ is the output projection. The MLP term operates on the mid-layer residual $h_t^{\text{mid},l} = \text{LayerNorm}(h_t^{(l-1)} + \text{MHA}^l_t)$:
\begin{equation}
    \text{MLP}^l_t = W^l_{\text{proj}} \; \sigma\!\left(W^l_{\text{fc}} \; h_t^{\text{mid},l}\right),
    \label{equation:mlp}
\end{equation}
where $\sigma$ denotes a non-linear activation function, $W^l_{\text{fc}} \in \mathbb{R}^{d_{\text{ff}} \times d_{\text{model}}}$ is the up-projection, and $W^l_{\text{proj}} \in \mathbb{R}^{d_{\text{model}} \times d_{\text{ff}}}$ is the down-projection. The final prediction is obtained by applying a language modelling head to the last-position hidden state $h_T^{(L)}$.

\subsection{ROME Formalisation}\label{appendix:rome}
Rank-One Model Editing (ROME)~\citep{meng2022locating} provides the causal tracing framework adapted in this work. ROME interprets MLP layers as implicit key--value stores. At layer $l$, the MLP output for token $t$ can be written as:
\begin{equation}
    \text{MLP}^l_t = W^l_{\text{proj}} \; \sigma\!\left(W^l_{\text{fc}} \; h_t^{\text{mid},l}\right) = W^l_{\text{proj}} \; k_t^l,
    \label{equation:rome_kv}
\end{equation}
where $k_t^l = \sigma(W^l_{\text{fc}} \; h_t^{\text{mid},l}) \in \mathbb{R}^{d_{\text{ff}}}$ serves as the key vector and each column of $W^l_{\text{proj}}$ serves as a value vector. Under this interpretation, the MLP computes a weighted combination of stored values indexed by the input-dependent key. ROME exploits this structure to locate factual associations by identifying the layer $l^*$ and subject token $t^*$ at which restoring clean-run activations maximally recovers the correct prediction. The present work adapts the localisation procedure to the reverse dictionary setting; model editing is not performed.

\subsection{Causal Tracing Protocol}\label{appendix:causal_tracing}
The causal tracing procedure follows \citet{meng2022locating} and proceeds in three stages.

\paragraph{Stage 1: Clean execution.}
The model processes the original input $X$ and caches internal states $\{h_t^{(l)}, \text{MHA}^l_t, \text{MLP}^l_t\}$ for all layers $l$ and token positions $t$. The clean-run probability assigned to the correct definiendum token $y$ is recorded as $p_y(X)$.

\paragraph{Stage 2: Corrupted execution.}
Gaussian noise $\mathcal{N}(0, (3\sigma_e)^2)$ is added to the token embeddings of the definition span only, where $\sigma_e$ is the empirical standard deviation of embedding activations computed over the dataset. This produces a corrupted input $X^*$ and a degraded prediction probability $p_y(X^*)$. Each input is evaluated using one clean run and 10 corrupted runs with independent noise samples.

\paragraph{Stage 3: Restoration (patching).}
For each component type $k \in \{\text{MLP}, \text{MHA}, \text{hidden}\}$, the corrupted activation at a target position $(i, j)$ is replaced with the corresponding clean-run value, yielding $p_y(X^{*,\text{patch}(i,j,k)})$. This patched probability is used to compute the normalised recovery score in Eq.~\ref{eq:recovery}. For MLP and MHA components, restoration is applied over a sliding window of 10 consecutive layers centred at layer index $j$, matching the protocol of \citet{meng2022locating}. For hidden states, restoration targets individual layer--token pairs.

Recovery scores are averaged across the 10 corrupted runs to obtain stable estimates. Quantitative summaries are based on the top-$K$ restored states per example, defined as the $(i, j)$ pairs with the largest recovery scores $\mathbf{T}^{(k)}_{ij}$ (Eq.~\ref{eq:recovery}). Unless stated otherwise, $K = 50$ is used for reporting, with $K = 10$ provided for comparison.

\begin{figure}[t]
    \centering
    \begin{subfigure}{0.9\linewidth}
        \centering
        \includegraphics[width=\linewidth]{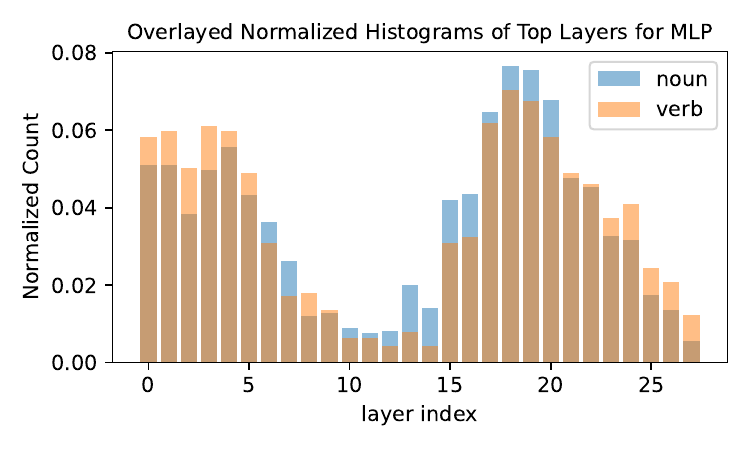}
        \caption{}
        \label{subfig:normalised_histograms_pos_mlp}
    \end{subfigure}
    \begin{subfigure}{0.9\linewidth}
        \centering
        \includegraphics[width=\linewidth]{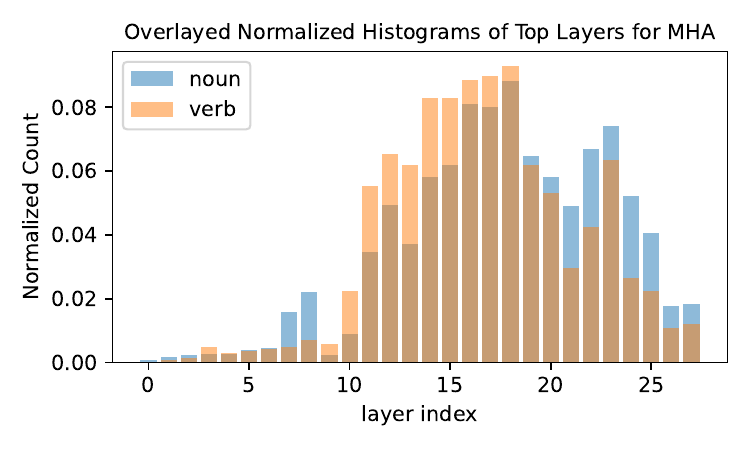}
        \caption{}
        \label{subfig:normalised_histograms_pos_attn}
    \end{subfigure}
    \begin{subfigure}{0.9\linewidth}
        \centering
        \includegraphics[width=\linewidth]{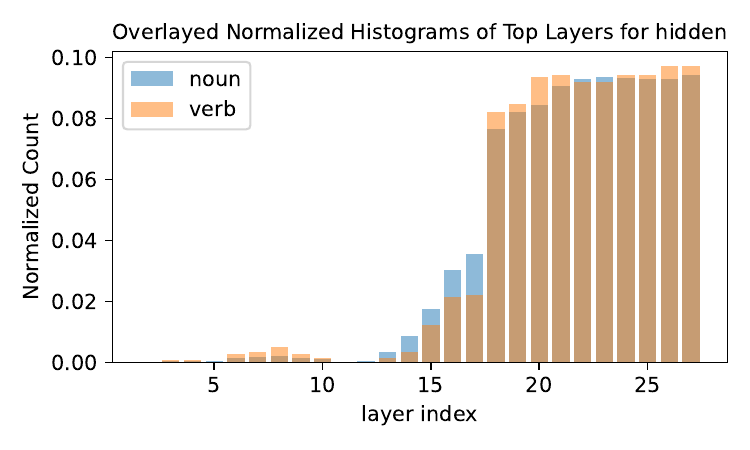}
        \caption{}
        \label{subfig:normalised_histograms_pos_hidden}
    \end{subfigure}
    \caption{Top-10 restored states from (a)~MLP, (b)~MHA, and (c)~hidden components, grouped by the part of speech of the definiendum (GPT-J-6B, EWN, $K = 10$). Normalised histograms are overlaid. X-axis: layer index; y-axis: normalised count.}
    \label{fig:normalised_histograms_pos}
\end{figure}

\begin{figure}[t]
    \centering
    \begin{subfigure}{\linewidth}
        \centering
        \includegraphics[width=\columnwidth]{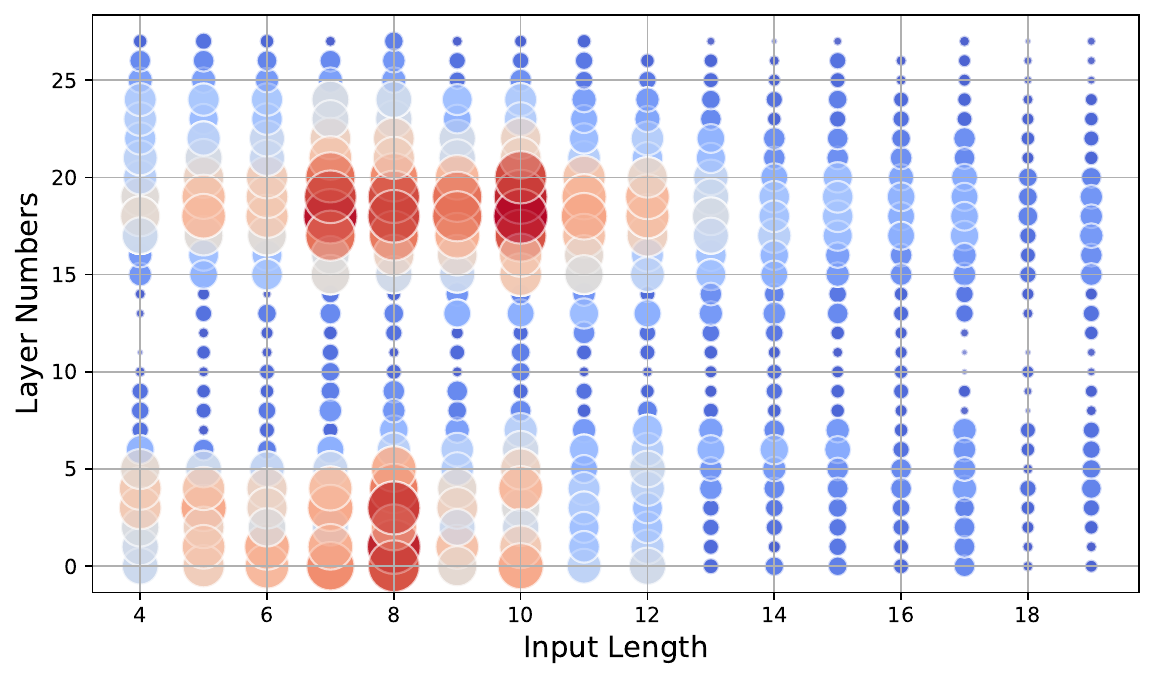}
        \caption{MLP.}
        \label{subfig:input_length_vs_top_states_mlp}
    \end{subfigure}
    \begin{subfigure}{\linewidth}
        \centering
        \includegraphics[width=\columnwidth]{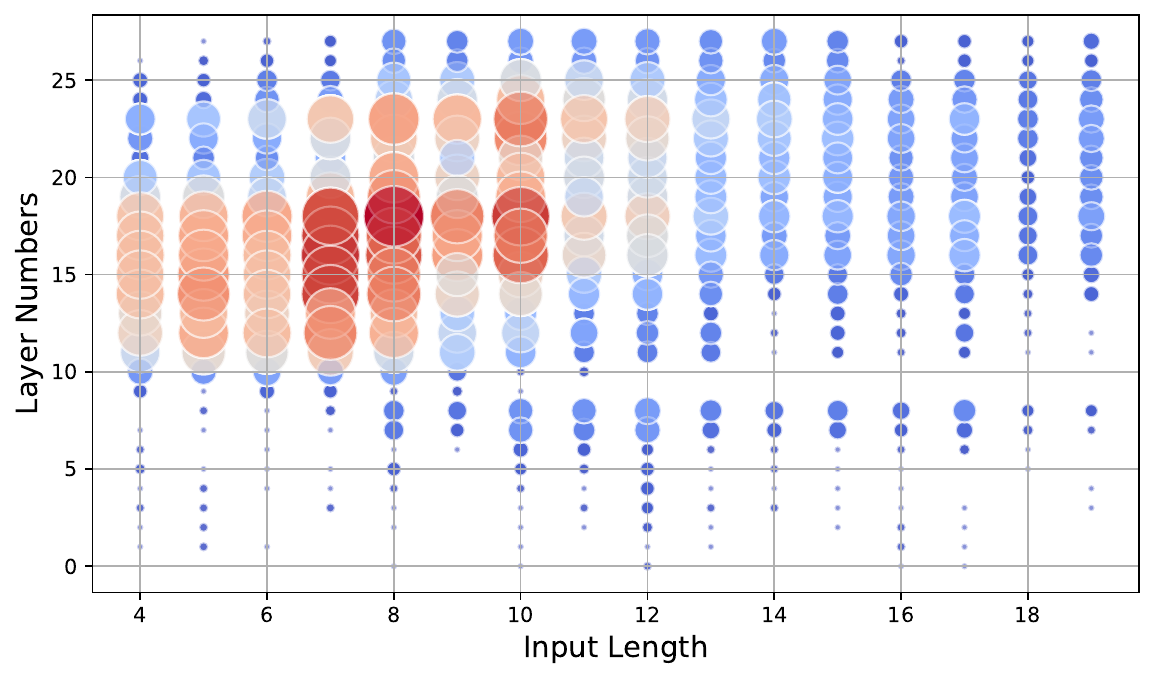}
        \caption{MHA.}
        \label{subfig:input_length_vs_top_states_attn}
    \end{subfigure}
    \begin{subfigure}{\linewidth}
        \centering
        \includegraphics[width=\columnwidth]{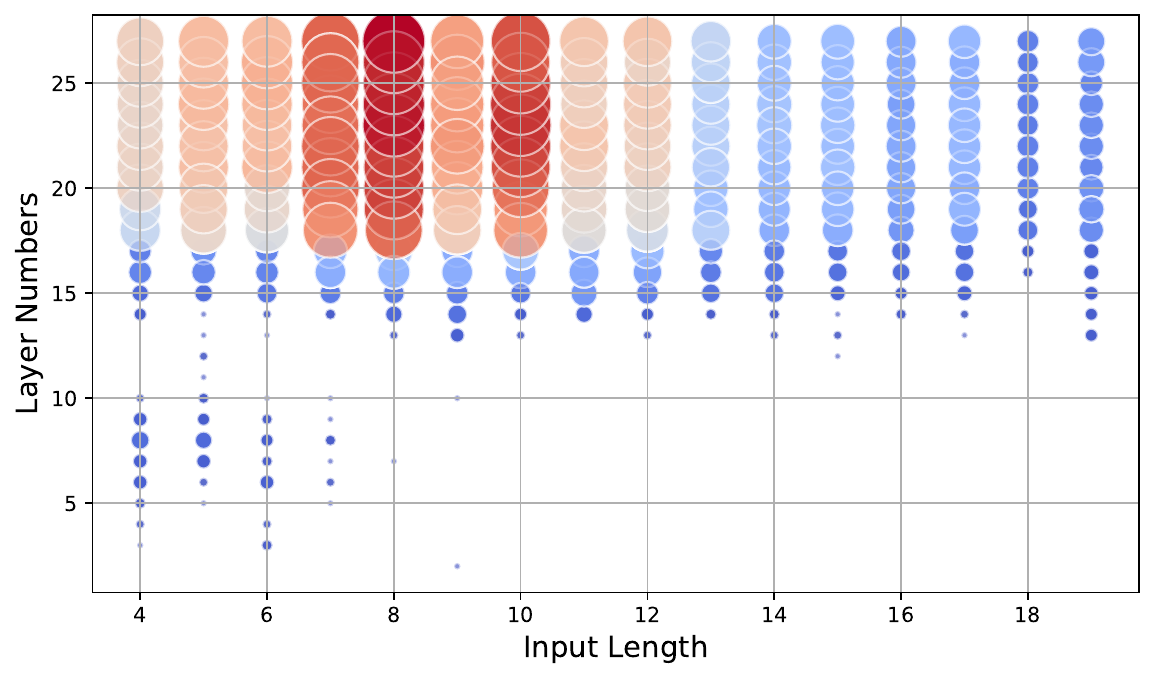}
        \caption{Hidden states.}
        \label{subfig:input_length_vs_top_states_hidden}
    \end{subfigure}
    \caption{Top-10 restored states from (a)~MLP, (b)~MHA, and (c)~hidden components plotted against input length (GPT-J-6B, EWN, $K = 10$). X-axis: input length (tokens); y-axis: layer index of top states. No systematic association between definition length and the layer distribution of restored states is observed.}
    \label{fig:input_length_vs_top_states}
\end{figure}

\section{Supplementary Results}\label{appendix:supplementary_results}

\subsection{PoS Grouping}\label{appendix:pos_grouping}
Figure~\ref{fig:normalised_histograms_pos} presents overlaid normalised histograms of top-10 restored states grouped by the part of speech of the definiendum (nouns vs.\ verbs). The distributions are broadly similar across all three component types, indicating that the PoS of the definiendum has minimal effect on localisation patterns. Minor differences are observable: in MLP layers, noun-related states are slightly more concentrated in upper layers compared to verbs; in MHA layers, the distribution for verbs approximates a Gaussian shape while that for nouns exhibits slight negative skew; no clear differences are discernible in hidden layers. These observations suggest that the localisation patterns reported in Section~4 are robust to variation in definiendum PoS and are therefore driven primarily by definitional content rather than the grammatical category of the target word.

\subsection{Correlation with Definition Length}\label{appendix:length_correlation}
Figure~\ref{fig:input_length_vs_top_states} plots the layer indices of top-10 restored states against input length for each component type. No systematic association between definition length and the layer distribution of influential states is observed across MLP, MHA, or hidden components. This result indicates that the localisation patterns reported in Section~4 are not confounded by input length and that the models apply consistent processing strategies across definitions of varying complexity.

\subsection{DSR-Labelled MHA and Hidden-State Traces}\label{appendix:dsr_traces}
\begin{figure*}[h!]
    \begin{subfigure}{0.33\linewidth}
        \centering
        \includegraphics[width=\linewidth]{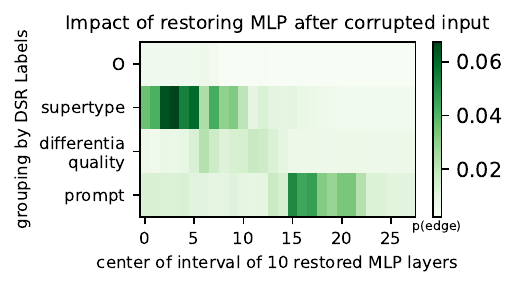}
        \caption{}
        \label{subfig:DSR_MLP_1}
    \end{subfigure}
    \begin{subfigure}{0.33\linewidth}
        \centering
        \includegraphics[width=\linewidth]{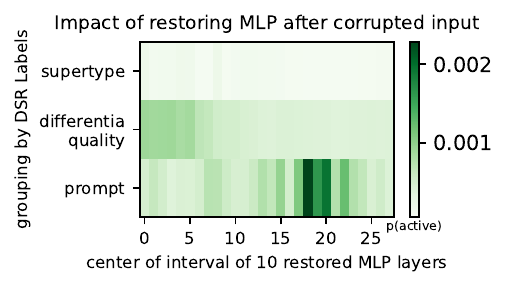}
        \caption{}
        \label{subfig:DSR_MLP_2}
    \end{subfigure}
    \begin{subfigure}{0.33\linewidth}
        \centering
        \includegraphics[width=\linewidth]{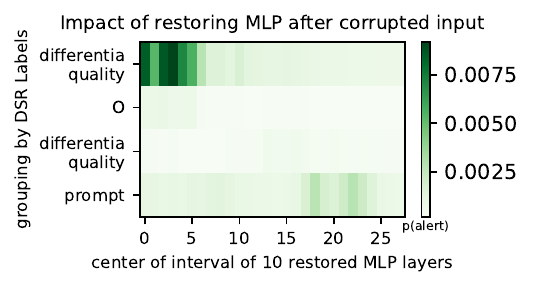}
        \caption{}
        \label{subfig:DSR_MLP_3}
    \end{subfigure}
    \caption{Sample of causal tracing with DSR labelling when restoring a window of 10 MLP layers. The representation highlights the distribution of important states over several layers and the importance of content words, mainly captured in supertype (a) and differentia quality (b) and (c).}  
    \label{fig:DSR_MLP} 
\end{figure*}

\begin{figure*}[t]
    \begin{subfigure}{0.33\linewidth}
        \centering
        \includegraphics[width=\linewidth]{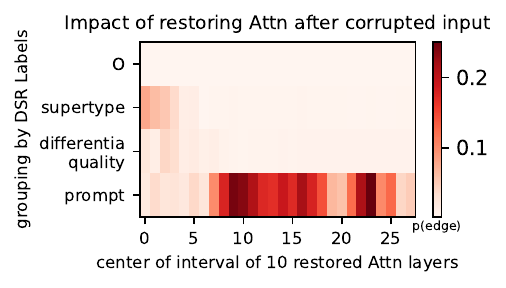}
        \caption{}
        \label{subfig:DSR_attn_1}
    \end{subfigure}
    \begin{subfigure}{0.33\linewidth}
        \centering
        \includegraphics[width=\linewidth]{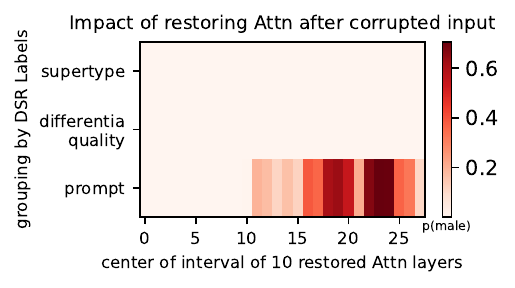}
        \caption{}
        \label{subfig:DSR_attn_2}
    \end{subfigure}
    \begin{subfigure}{0.33\linewidth}
        \centering
        \includegraphics[width=\linewidth]{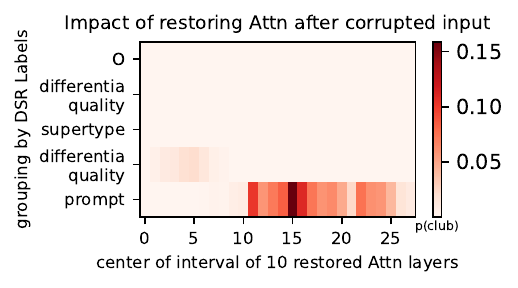}
        \caption{}
        \label{subfig:DSR_attn_3}
    \end{subfigure}
    \caption{Causal traces with DSR labelling for MHA restoration (GPT-J-6B, EWN). Each heatmap shows recovery probability $p_y$ (colour scale) when restoring a window of 10 MHA layers (x-axis: window centre) at tokens grouped by DSR label (y-axis). (a)~\textit{edge}, (b)~\textit{male}, (c)~\textit{club}.}
    \label{fig:DSR_attn}
\end{figure*}

\begin{figure*}[t]
    \begin{subfigure}{0.33\linewidth}
        \centering
        \includegraphics[width=\linewidth]{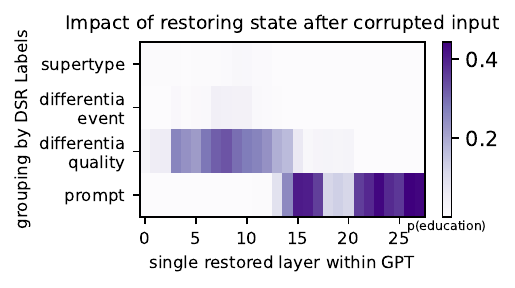}
        \caption{}
        \label{subfig:DSR_hidden_1}
    \end{subfigure}
    \begin{subfigure}{0.33\linewidth}
        \centering
        \includegraphics[width=\linewidth]{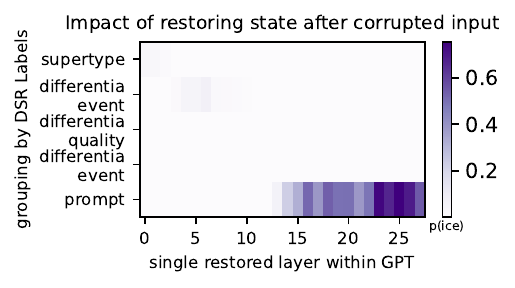}
        \caption{}
        \label{subfig:DSR_hidden_2}
    \end{subfigure}
    \begin{subfigure}{0.33\linewidth}
        \centering
        \includegraphics[width=\linewidth]{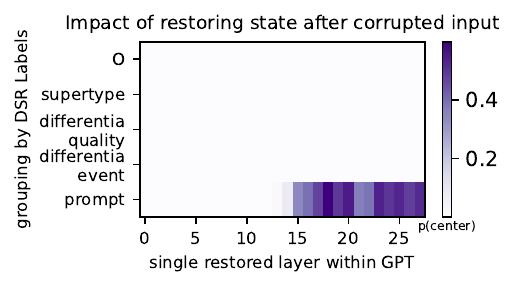}
        \caption{}
        \label{subfig:DSR_hidden_3}
    \end{subfigure}
    \caption{Causal traces with DSR labelling for hidden-state restoration (GPT-J-6B, EWN). Each heatmap shows recovery probability $p_y$ (colour scale) when restoring a single hidden state (x-axis: layer index) at tokens grouped by DSR label (y-axis). (a)~\textit{education}, (b)~\textit{ice}, (c)~\textit{center}.}
    \label{fig:DSR_hidden}
\end{figure*}

Figures~\ref{fig:DSR_MLP}, ~\ref{fig:DSR_attn} and~\ref{fig:DSR_hidden} show the DSR-grouped recovery patterns for MLP, MHA and hidden-state components respectively. For MHA restoration (Figure~\ref{fig:DSR_attn}), recovery is concentrated at prompt tokens across middle-to-upper layers, consistent with the aggregation pattern reported in Section~\ref{sec:empirical}. DSR-labelled definition tokens contribute comparatively low recovery at the top-10 threshold, but their contribution increases at broader aggregation levels (Table~1), indicating that MHA integrates definitional content progressively rather than at discrete layer--token positions. For hidden-state restoration (Figure~\ref{fig:DSR_hidden}), recovery is dominated by upper-layer states at the final token, with \textit{supertype} and differentia tokens exhibiting localised contributions in layers 20--27. These patterns are consistent with the upper-layer concentration reported in Section~\ref{sec:empirical} and support the observation (RQ3) that definitional semantic structure persists through aggregation into the final representation.

\subsection{Alternative Definitions}\label{appendix:alt_defs}
\begin{figure*}[t]
    \begin{subfigure}{0.32\linewidth}
        \includegraphics[width=\linewidth]{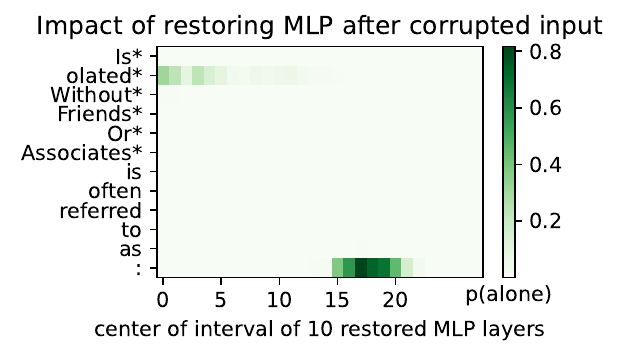}
        \caption{}
        \label{subfig:Alternative_definitions_MLP_1}
    \end{subfigure}
    \begin{subfigure}{0.32\linewidth}
        \includegraphics[width=\linewidth]{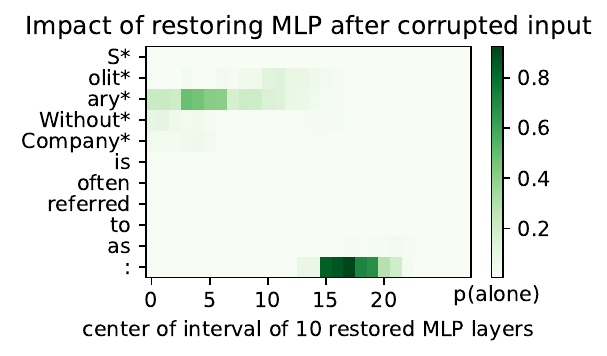}
        \caption{}
        \label{subfig:Alternative_definitions_MLP_2}
    \end{subfigure}
    \begin{subfigure}{0.32\linewidth}
        \includegraphics[width=\linewidth]{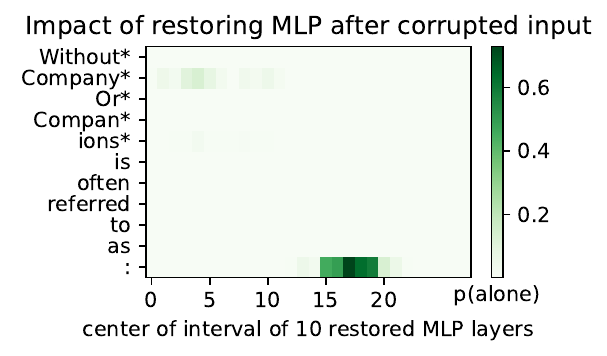}
        \caption{}
        \label{subfig:Alternative_definitions_MLP_3}
    \end{subfigure}
    \caption{MLP traces for alternative definitions of \textit{alone} (GPT-J-6B, EWN). Recovery probability $p_y$ (colour scale) when restoring a window of 10 MLP layers (x-axis: window centre) at each token (y-axis). Broad localisation patterns are preserved across paraphrases; specific high-impact positions shift.}
    \label{fig:Alternative_definitions_MLP}
\end{figure*}

\begin{figure*}[t]
    \begin{subfigure}{0.32\linewidth}
        \includegraphics[width=\linewidth]{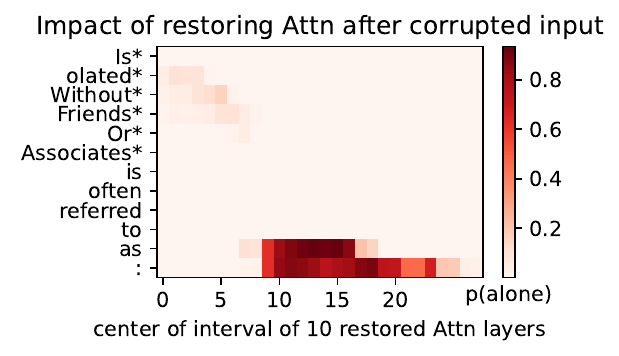}
        \caption{}
        \label{subfig:Alternative_definitions_attention_1}
    \end{subfigure}
    \begin{subfigure}{0.32\linewidth}
        \includegraphics[width=\linewidth]{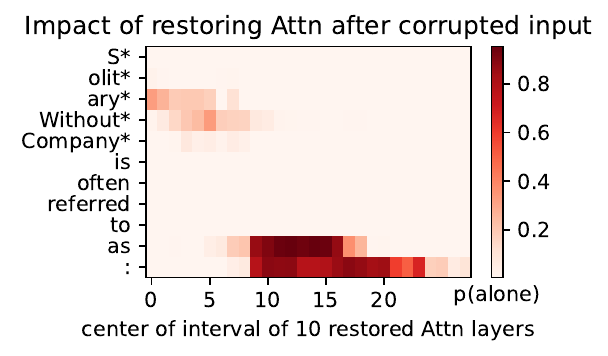}
        \caption{}
        \label{subfig:Alternative_definitions_attention_2}
    \end{subfigure}
    \begin{subfigure}{0.32\linewidth}
        \includegraphics[width=\linewidth]{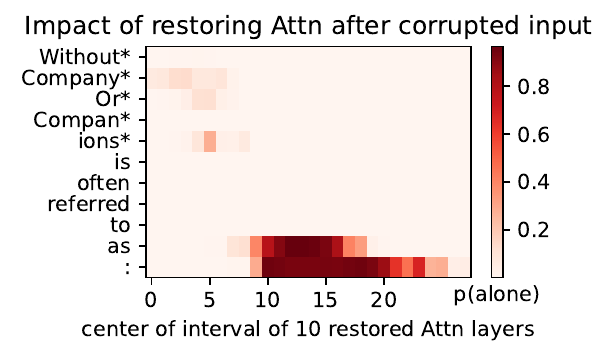}
        \caption{}
        \label{subfig:Alternative_definitions_attention_3}
    \end{subfigure}
    \caption{MHA traces for alternative definitions of \textit{alone} (GPT-J-6B, EWN). Recovery probability $p_y$ (colour scale) when restoring a window of 10 MHA layers (x-axis: window centre) at each token (y-axis). MHA patterns are more stable across paraphrases than MLP patterns.}
    \label{fig:Alternative_definitions_attention}
\end{figure*}

\begin{figure*}[t]
    \begin{subfigure}{0.33\linewidth}
        \includegraphics[width=\linewidth]{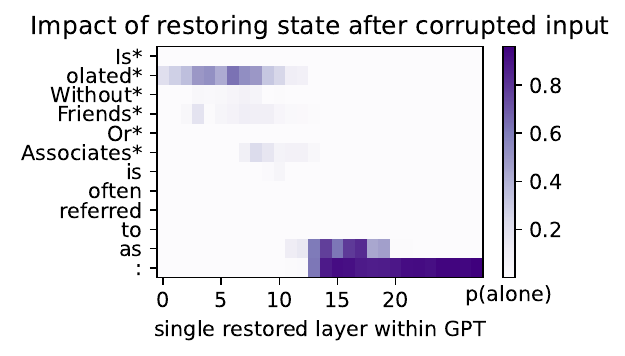}
        \caption{}
        \label{subfig:Alternative_definitions_hidden_1}
    \end{subfigure}
    \begin{subfigure}{0.33\linewidth}
        \includegraphics[width=\linewidth]{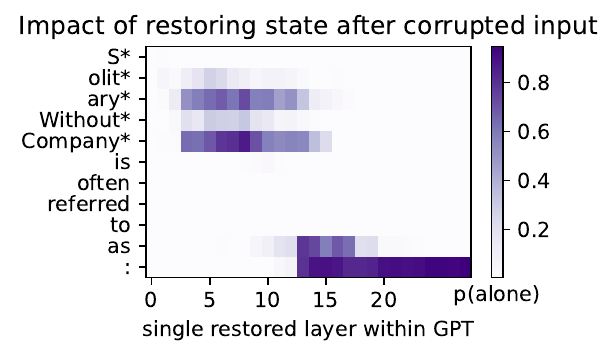}
        \caption{}
        \label{subfig:Alternative_definitions_hidden_2}
    \end{subfigure}
    \begin{subfigure}{0.33\linewidth}
        \includegraphics[width=\linewidth]{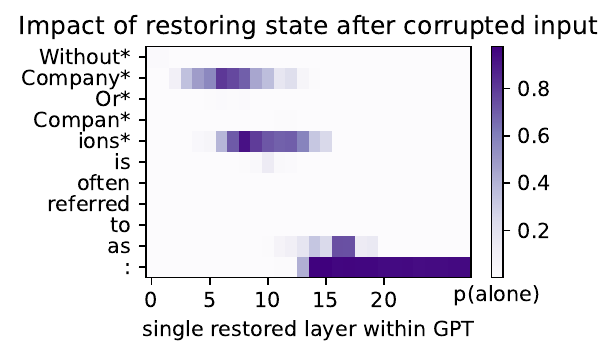}
        \caption{}
        \label{subfig:Alternative_definitions_hidden_3}
    \end{subfigure}
    \caption{Hidden-state traces for alternative definitions of \textit{alone} (GPT-J-6B, EWN). Recovery probability $p_y$ (colour scale) when restoring a single hidden state (x-axis: layer index) at each token (y-axis). Upper-layer concentration at the last token is consistent across paraphrases.}
    \label{fig:Alternative_definitions_hidden}
\end{figure*}

Figures~\ref{fig:Alternative_definitions_MLP}--\ref{fig:Alternative_definitions_hidden} present causal traces for three alternative definitions of the definiendum \textit{alone}, generated as described in Appendix~\ref{appendix:alt_defs_generation}. These traces serve as a robustness check: if localisation patterns are artefacts of specific surface forms, they should vary substantially across semantically equivalent paraphrases.

For MLP restoration (Figure~\ref{fig:Alternative_definitions_MLP}), the broad bimodal pattern (early and upper layers) is preserved across all three alternatives, although the specific token positions associated with high recovery shift in accordance with lexical variation. This is consistent with MLP layers encoding content-specific associations that depend on token identity while maintaining a stable layer-level distribution. For MHA restoration (Figure~\ref{fig:Alternative_definitions_attention}), patterns are more stable across paraphrases, with recovery consistently concentrated at the last token in middle-to-upper layers regardless of surface form. For hidden-state restoration (Figure~\ref{fig:Alternative_definitions_hidden}), upper-layer concentration at the final token is consistent across all three alternatives, confirming that this pattern reflects a structural property of the aggregation process rather than a lexical artefact.

\subsection{Supplementary English Results}\label{appendix:supplementary_results_en}
\begin{figure*}[t]
    \centering
    \begin{subfigure}{0.32\textwidth}
        \includegraphics[width=\linewidth]{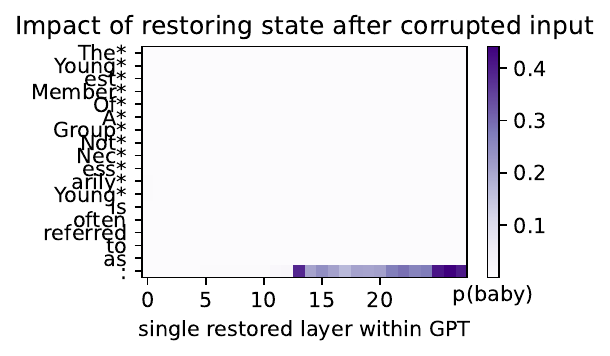}
        \caption{} \label{fig:restored_hidden_1_appendix}
    \end{subfigure}
    \hfill
    \begin{subfigure}{0.32\textwidth}
        \includegraphics[width=\linewidth]{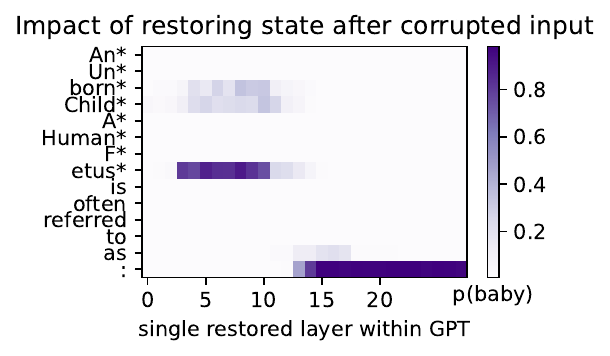}
        \caption{} \label{fig:restored_hidden_9_appendix}
    \end{subfigure}
    \hfill
    \begin{subfigure}{0.32\textwidth}
        \includegraphics[width=\linewidth]{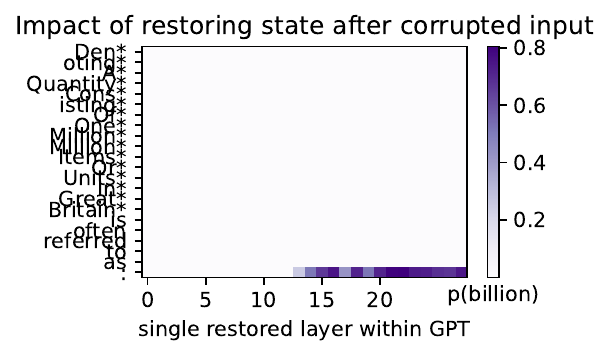}
        \caption{} \label{fig:restored_hidden_11_appendix}
    \end{subfigure}
    \begin{subfigure}{0.32\textwidth}
        \includegraphics[width=\linewidth]{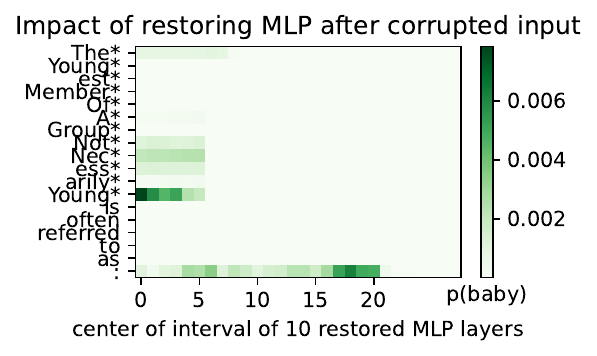}
        \caption{} \label{fig:restored_mlp_1_appendix}
    \end{subfigure}
    \hfill
    \begin{subfigure}{0.32\textwidth}
        \includegraphics[width=\linewidth]{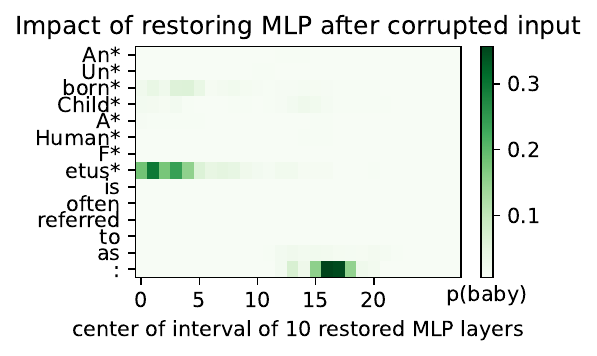}
        \caption{} \label{fig:restored_mlp_9_appendix}
    \end{subfigure}
    \hfill
    \begin{subfigure}{0.32\textwidth}
        \includegraphics[width=\linewidth]{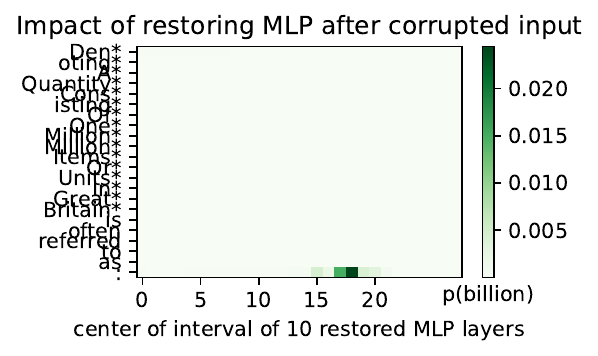}
        \caption{} \label{fig:restored_mlp_11_appendix}
    \end{subfigure}
    \begin{subfigure}{0.32\textwidth}
        \includegraphics[width=\linewidth]{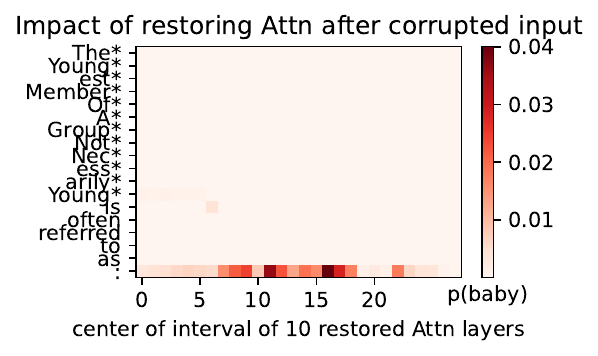}
        \caption{} \label{fig:restored_attn_1_appendix}
    \end{subfigure}
    \hfill
    \begin{subfigure}{0.32\textwidth}
        \includegraphics[width=\linewidth]{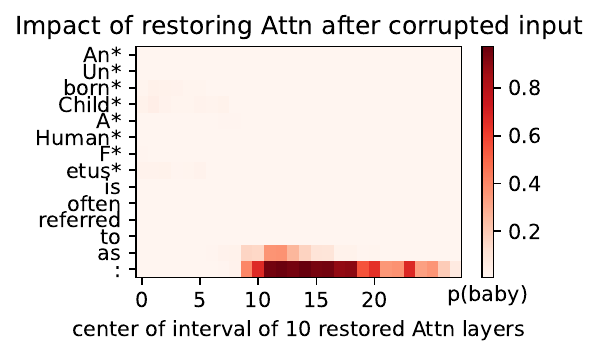}
        \caption{} \label{fig:restored_attn_9_appendix}
    \end{subfigure}
    \hfill
    \begin{subfigure}{0.32\textwidth}
        \includegraphics[width=\linewidth]{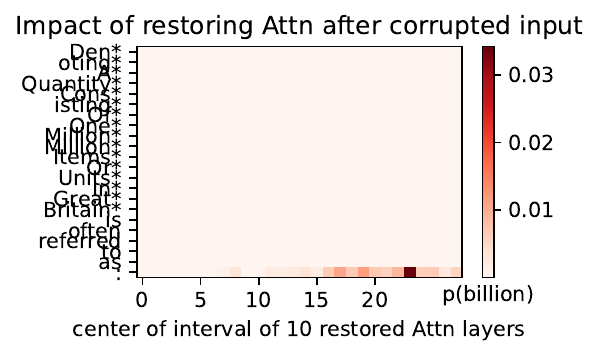}
        \caption{} \label{fig:restored_attn_11_appendix}
    \end{subfigure}
    \caption{Supplementary causal traces for GPT-J-6B (EWN). Top row: hidden-state restoration (single layer); middle row: MLP restoration (window of 10 layers); bottom row: MHA restoration (window of 10 layers). Columns show different definienda: (a,d,g)~\textit{baby} (definition 1); (b,e,h)~\textit{baby} (definition 2); (c,f,i)~\textit{billion}. Colour scale: recovery probability $p_y$, clipped to $[0,1]$.}
    \label{fig:combined_states_appendix_en}
\end{figure*}

Figure~\ref{fig:combined_states_appendix_en} presents additional causal tracing results that illustrate two phenomena: homonym handling and numerical concept processing.

Figures~\ref{fig:restored_hidden_1_appendix} and~\ref{fig:restored_hidden_9_appendix} show traces for two distinct definitions of the definiendum \textit{baby}. For the second definition, which contains words strongly linked to the definiendum (e.g., \textit{child}, \textit{fetus}), hidden-state recovery is concentrated entirely at the last token (Figure~\ref{fig:restored_hidden_9_appendix}). In contrast, for the first definition, which lacks such direct lexical cues, additional definition tokens participate in high-recovery states (Figure~\ref{fig:restored_hidden_1_appendix}). MLP layers exhibit greater sensitivity to semantically related words: when the definition contains tokens with strong lexical association to the definiendum, multiple early-layer token positions contribute to recovery (Figure~\ref{fig:restored_mlp_9_appendix}). MHA layers maintain consistent last-token concentration across both definitions (Figures~\ref{fig:restored_attn_1_appendix} and~\ref{fig:restored_attn_9_appendix}), further supporting the component differentiation reported in Section~4 (RQ2).

Number handling provides an additional test case. For the definiendum \textit{billion} (Figures~\ref{fig:restored_hidden_11_appendix} and~\ref{fig:restored_mlp_11_appendix}), recovery is not associated with tokens containing other numerical values (e.g., \textit{million}, \textit{one}). Instead, the model treats spelled-out numbers as conceptual units rather than attending to their arithmetic content, consistent with the compositional interpretation account in which meaning is derived from definitional structure rather than token-level numerical matching.

\subsection{Supplementary Spanish Results}\label{appendix:supplementary_results_es}
\begin{figure*}[t]
    \centering
    \begin{subfigure}[t]{0.32\linewidth}
        \centering
        \includegraphics[width=\linewidth]{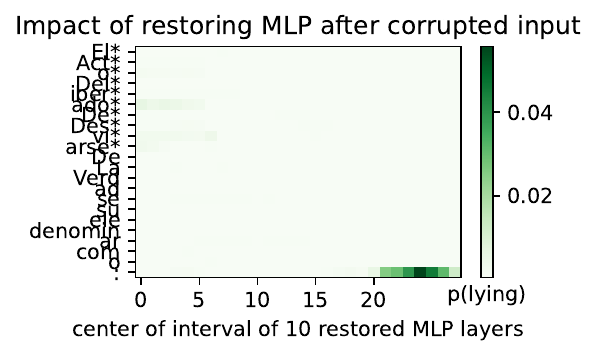}
        \caption{}
        \label{subfig:spanish_mlp_1}
    \end{subfigure}
    \begin{subfigure}[t]{0.32\linewidth}
        \centering
        \includegraphics[width=\linewidth]{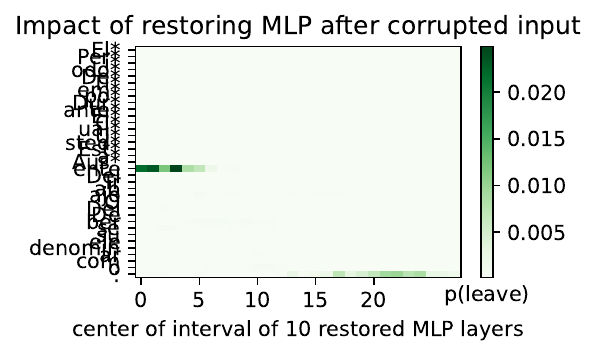}
        \caption{}
        \label{subfig:spanish_mlp_2}
    \end{subfigure}
    \begin{subfigure}[t]{0.32\linewidth}
        \centering
        \includegraphics[width=\linewidth]{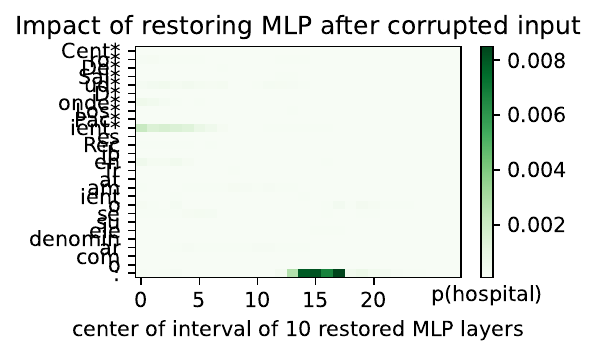}
        \caption{}
        \label{subfig:spanish_mlp_3}
    \end{subfigure}
    \begin{subfigure}[t]{0.32\linewidth}
        \centering
        \includegraphics[width=\linewidth]{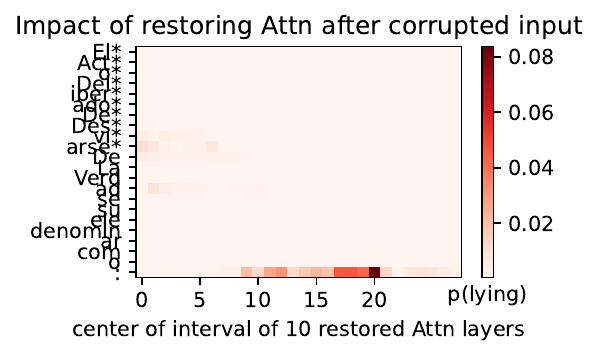}
        \caption{}
        \label{subfig:spanish_attention_1}
    \end{subfigure}
    \begin{subfigure}[t]{0.32\linewidth}
        \centering
        \includegraphics[width=\linewidth]{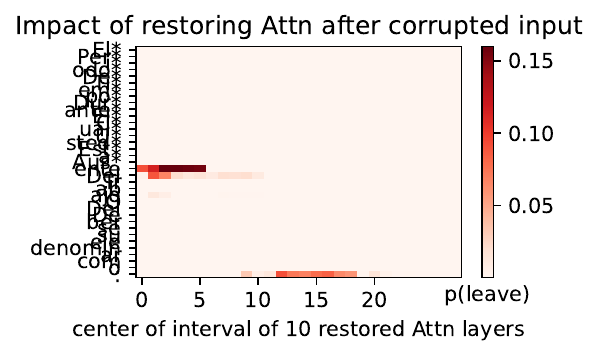}
        \caption{}
        \label{subfig:spanish_attention_2}
    \end{subfigure}
    \begin{subfigure}[t]{0.32\linewidth}
        \centering
        \includegraphics[width=\linewidth]{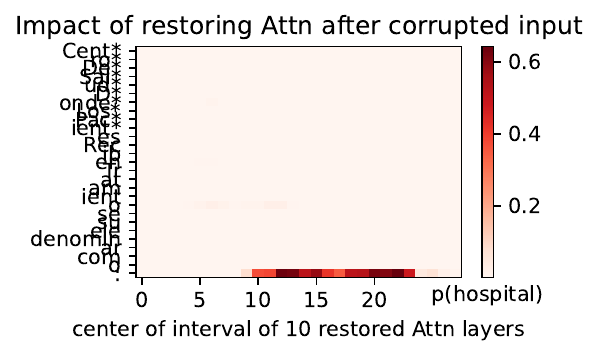}
        \caption{}
        \label{subfig:spanish_attention_3}
    \end{subfigure}
    \begin{subfigure}[t]{0.32\linewidth}
        \centering
        \includegraphics[width=\linewidth]{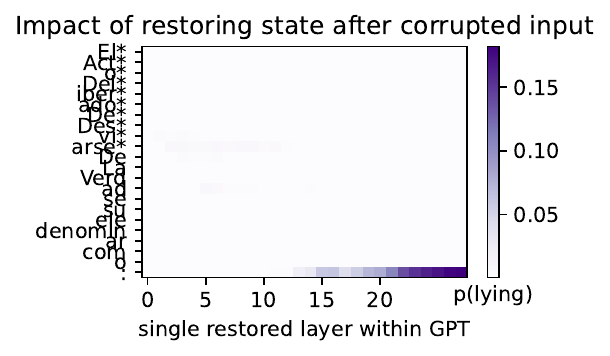}
        \caption{}
        \label{subfig:spanish_hidden_states_1}
    \end{subfigure}
    \begin{subfigure}[t]{0.32\linewidth}
        \centering
        \includegraphics[width=\linewidth]{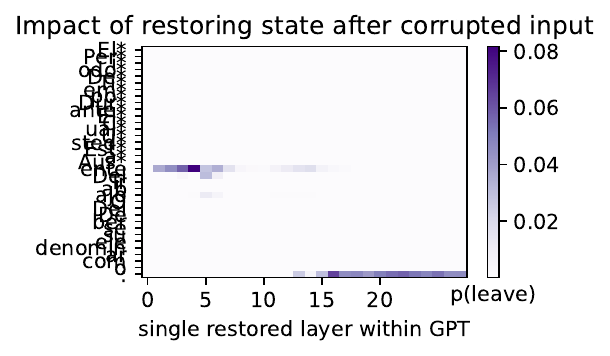}
        \caption{}
        \label{subfig:spanish_hidden_states_2}
    \end{subfigure}
    \begin{subfigure}[t]{0.32\linewidth}
        \centering
        \includegraphics[width=\linewidth]{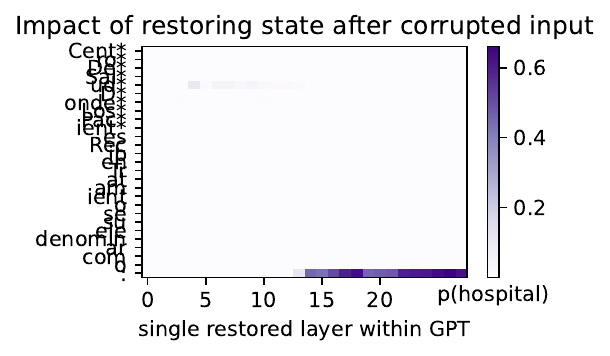}
        \caption{}
        \label{subfig:spanish_hidden_states_3}
    \end{subfigure}
    \caption{Cross-lingual causal traces for BERTIN GPT-J-6B (SWN). Top row: MLP restoration (window of 10 layers); middle row: MHA restoration (window of 10 layers); bottom row: hidden-state restoration (single layer). Columns show different definienda: (a,d,g)~\textit{lying}; (b,e,h)~\textit{leave}; (c,f,i)~\textit{hospital}. Colour scale: recovery probability $p_y$, clipped to $[0,1]$. Y-axis: input tokens (asterisks denote corrupted definition tokens). X-axis: window centre (MLP, MHA) or layer index (hidden states).}
    \label{fig:spanish_combined_traces}
\end{figure*}

Figure~\ref{fig:spanish_combined_traces} presents causal traces for three definienda (\textit{lying}, \textit{leave}, \textit{hospital}) from SWN using BERTIN GPT-J-6B. These results replicate the English analysis and provide evidence for the cross-lingual stability of localisation patterns reported in Section~4.4 (RQ1).

MLP traces (Figure~\ref{fig:spanish_combined_traces}, a--c) exhibit early-layer recovery at content-bearing tokens, consistent with the bimodal pattern observed for GPT-J-6B on EWN. The distribution of high-impact positions is slightly broader than in the English setting, which is consistent with increased subword fragmentation in the Spanish tokenisation (e.g., \textit{Deliberado} $\rightarrow$ \textit{Del*}, \textit{iber*}, \textit{ado*}). MHA traces (Figure~\ref{fig:spanish_combined_traces}, d--f) show last-token concentration in middle-to-upper layers, mirroring the English pattern with no qualitative differences. Hidden-state traces (Figure~\ref{fig:spanish_combined_traces}, g--i) confirm upper-layer concentration at the final token across all three examples.

The consistency of these patterns across languages and tokenisation schemes supports the interpretation that the traced behaviours reflect architectural properties of the transformer components rather than artefacts of English-specific lexical structure.

\subsection{Layer-Index Distributions}\label{appendix:layer_distributions}
This subsection reports layer-index distributions of top-$K$ restored states for each component. The plots complement the example-level causal traces shown in Section~\ref{sec:empirical} by providing aggregated evidence for the bimodal MLP pattern and the unimodal MHA and hidden-state patterns.



\begin{figure*}[t]
\centering
\begin{subfigure}[t]{0.32\textwidth}
  \includegraphics[width=\linewidth]{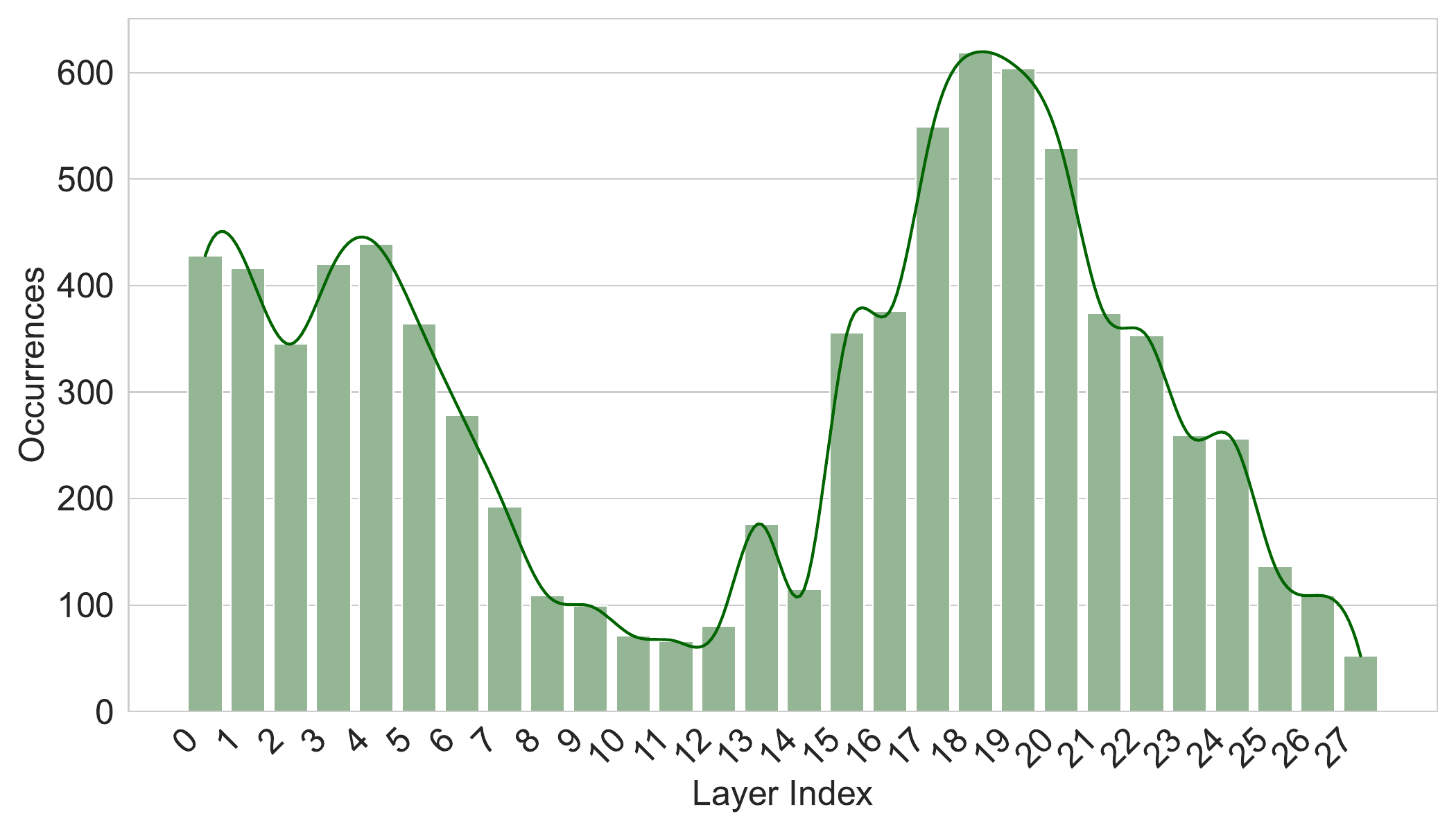}
  \caption{MLP}
  \label{fig:summary_mlp_plot_with_smooth_line}
\end{subfigure}
\hfill
\begin{subfigure}[t]{0.32\textwidth}
  \includegraphics[width=\linewidth]{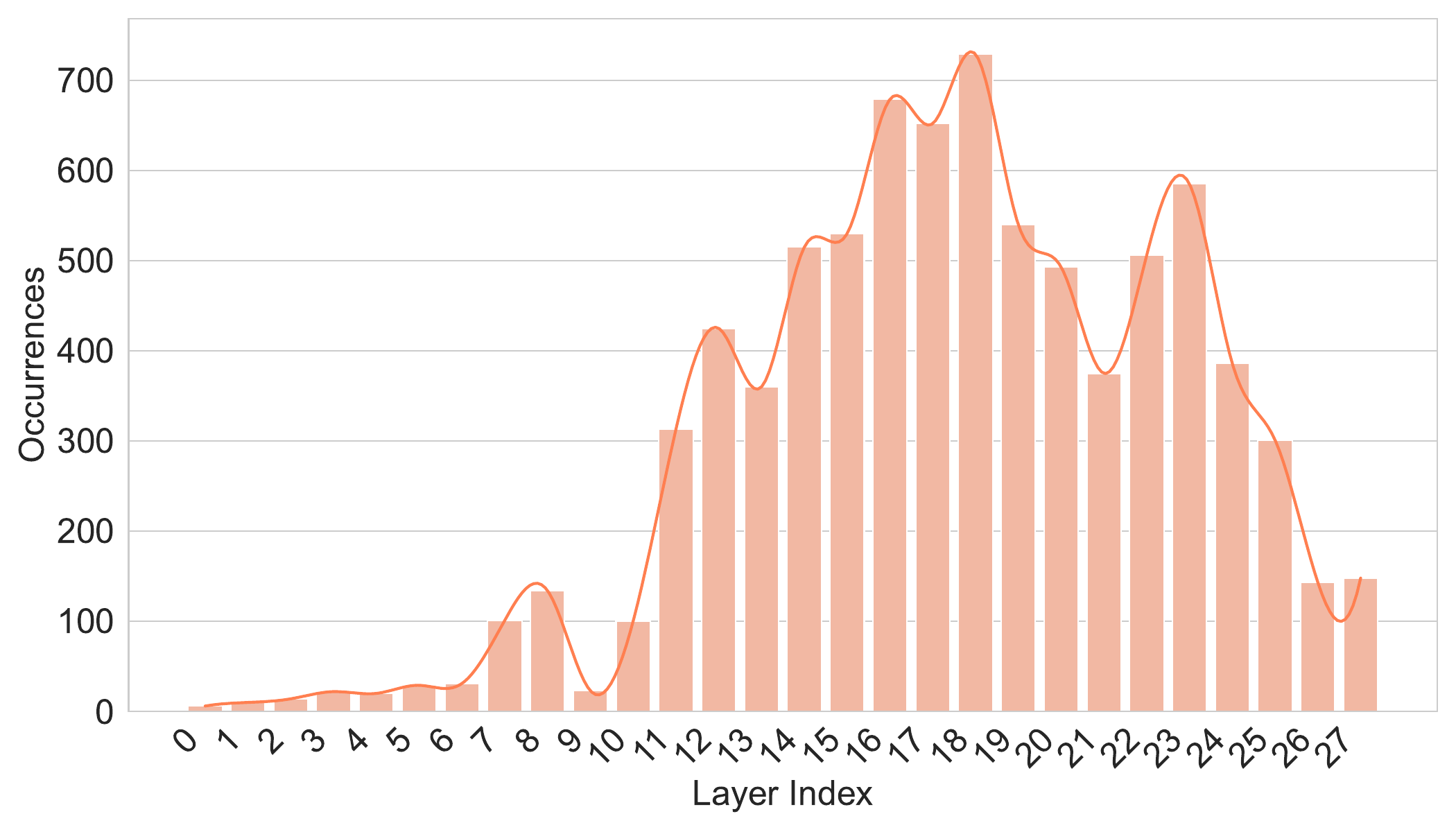}
  \caption{MHA}
  \label{fig:summary_atten_plot_with_smooth_line}
\end{subfigure}
\hfill
\begin{subfigure}[t]{0.32\textwidth}
  \includegraphics[width=\linewidth]{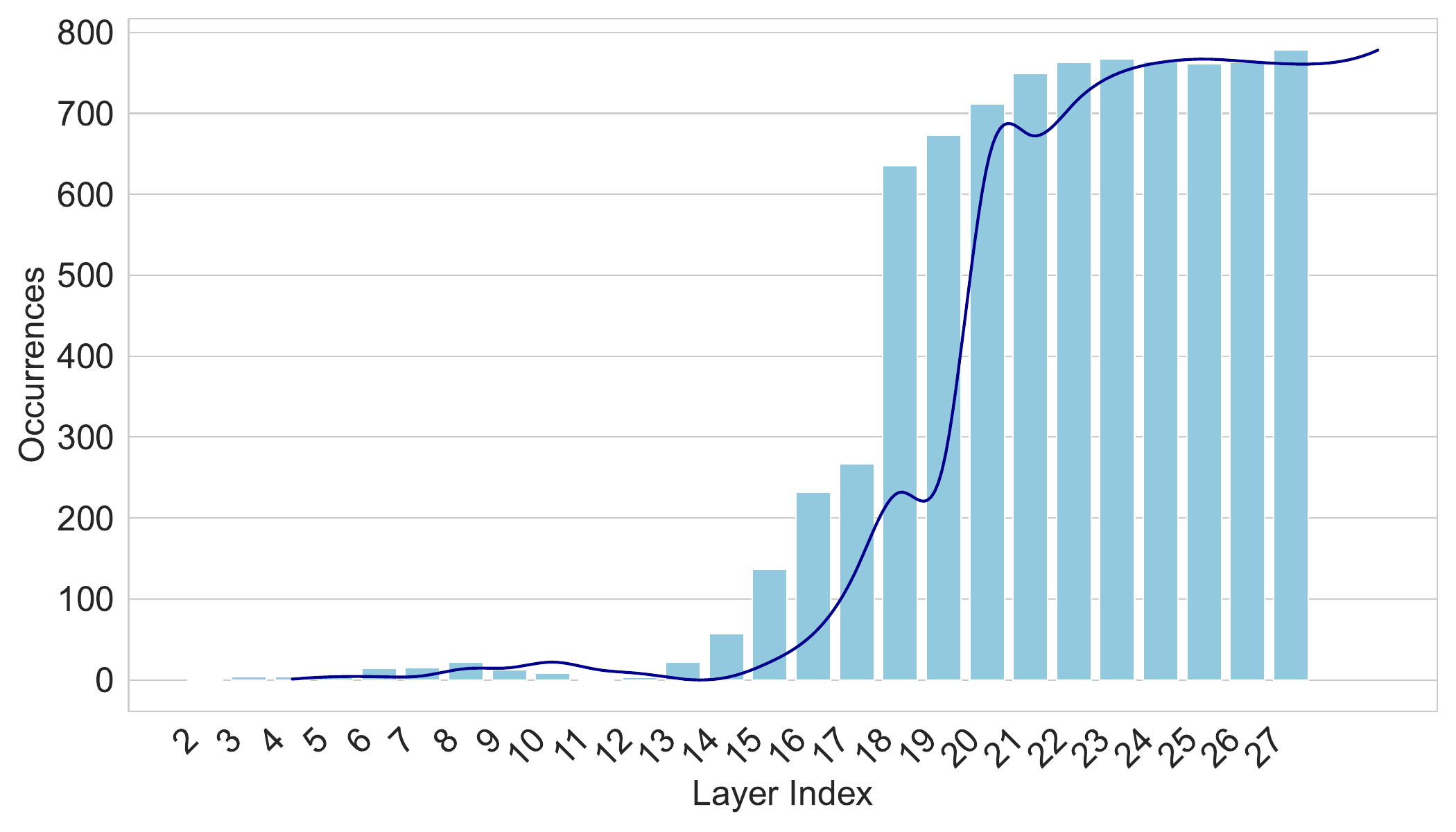}
  \caption{Hidden states}
  \label{fig:summary_hidden_plot_with_smooth_line}
\end{subfigure}

\caption{Distribution of layer indices among the top-10 restored states across correctly predicted GPT-J-6B EWN test samples. MLP restoration shows a bimodal pattern with clusters in early layers 0--5 and upper layers 17--21; MHA restoration concentrates in middle-to-upper layers 12--25; hidden-state restoration concentrates in upper layers 20--27.}
\label{fig:summary_restored_states_combined}
\end{figure*}

%% file: acl_latex.bib
@inproceedings{huben2023sparse,
  title={Sparse autoencoders find highly interpretable features in language models},
  author={Huben, Robert and Cunningham, Hoagy and Smith, Logan Riggs and Ewart, Aidan and Sharkey, Lee},
  booktitle={The Twelfth International Conference on Learning Representations},
  year={2023}
}

@inproceedings{
xiong2026toward,
title={Toward Faithful Retrieval-Augmented Generation with Sparse Autoencoders},
author={Guangzhi Xiong and Zhenghao He and Bohan Liu and Sanchit Sinha and Aidong Zhang},
booktitle={The Fourteenth International Conference on Learning Representations},
year={2026},
url={https://openreview.net/forum?id=hgBZP67BkP}
}

@inproceedings{
song2025llm,
title={{LLM} Interpretability with Identifiable Temporal-Instantaneous Representation},
author={Xiangchen Song and Jiaqi Sun and Zijian Li and Yujia Zheng and Kun Zhang},
booktitle={The Thirty-ninth Annual Conference on Neural Information Processing Systems},
year={2025},
url={https://openreview.net/forum?id=TdmzrkdLG0}
}

@article{tahimic2025mechanistic,
  title={Mechanistic Interpretability of Code Correctness in LLMs via Sparse Autoencoders},
  author={Tahimic, Kriz and Cheng, Charibeth},
  journal={arXiv preprint arXiv:2510.02917},
  year={2025}
}

@inproceedings{shu-etal-2025-survey,
    title = "A Survey on Sparse Autoencoders: Interpreting the Internal Mechanisms of Large Language Models",
    author = "Shu, Dong  and
      Wu, Xuansheng  and
      Zhao, Haiyan  and
      Rai, Daking  and
      Yao, Ziyu  and
      Liu, Ninghao  and
      Du, Mengnan",
    editor = "Christodoulopoulos, Christos  and
      Chakraborty, Tanmoy  and
      Rose, Carolyn  and
      Peng, Violet",
    booktitle = "Findings of the Association for Computational Linguistics: EMNLP 2025",
    month = nov,
    year = "2025",
    address = "Suzhou, China",
    publisher = "Association for Computational Linguistics",
    url = "https://aclanthology.org/2025.findings-emnlp.89/",
    doi = "10.18653/v1/2025.findings-emnlp.89",
    pages = "1690--1712",
    ISBN = "979-8-89176-335-7",
}

@article{meng2022locating,
  title={Locating and editing factual associations in GPT},
  author={Meng, Kevin and Bau, David and Andonian, Alex and Belinkov, Yonatan},
  journal={Advances in Neural Information Processing Systems},
  volume={35},
  pages={17359--17372},
  year={2022}
}

@book{jackendoff1992semantic,
  title={Semantic structures},
  author={Jackendoff, Ray S},
  volume={18},
  year={1992},
  publisher={MIT press}
}

@book{levin1993english,
  title={English verb classes and alternations: A preliminary investigation},
  author={Levin, Beth},
  year={1993},
  publisher={University of Chicago press}
}

@article{rappaport2008english,
  title={The English dative alternation: The case for verb sensitivityl},
  author={Rappaport Hovav, Malka and Levin, Beth},
  journal={Journal of linguistics},
  volume={44},
  number={1},
  pages={129--167},
  year={2008}
}

@book{smolensky2006harmonic,
  title={The harmonic mind: From neural computation to optimality-theoretic grammar (Cognitive architecture), Vol. 1},
  author={Smolensky, Paul and Legendre, G{\'e}raldine},
  year={2006},
  publisher={MIT press}
}

@inproceedings{clark2008compositional,
  title={A compositional distributional model of meaning},
  author={Clark, Stephen and Coecke, Bob and Sadrzadeh, Mehrnoosh},
  booktitle={Proceedings of the Second Quantum Interaction Symposium (QI-2008)},
  pages={133--140},
  year={2008},
  organization={Oxford}
}

@article{vaswani2017attention,
  title={Attention is all you need},
  author={Vaswani, Ashish and Shazeer, Noam and Parmar, Niki and Uszkoreit, Jakob and Jones, Llion and Gomez, Aidan N and Kaiser, {\L}ukasz and Polosukhin, Illia},
  journal={Advances in neural information processing systems},
  volume={30},
  year={2017}
}

@InProceedings{Gonzalez-Agirre:Laparra:Rigau:2012,
  author = "Aitor Gonzalez-Agirre and Egoitz Laparra and German Rigau",
  title = "Multilingual Central Repository version 3.0: upgrading a very large lexical knowledge base",
  booktitle = "Proceedings of the 6th Global WordNet Conference (GWC 2012)",
  year = 2012,
  address = "Matsue",
}

@misc{gpt-j,
  author = {Wang, Ben and Komatsuzaki, Aran},
  title = {{GPT-J-6B: A 6 Billion Parameter Autoregressive Language Model}},
  howpublished = {\url{https://github.com/kingoflolz/mesh-transformer-jax}},
  year = 2021,
  month = May
}

@inproceedings{BERTIN-GPT,
  author        = {Javier De la Rosa and Andres Fernández},
  editor        = {Manuel Montes-y-Gómez and Julio Gonzalo and Francisco Rangel and Marco Casavantes and Miguel Ángel Álvarez-Carmona and Gemma Bel-Enguix and Hugo Jair Escalante and Larissa Freitas and Antonio Miranda-Escalada and Francisco Rodríguez-Sánchez and Aiala Rosá and Marco Antonio Sobrevilla-Cabezudo and Mariona Taulé and Rafael Valencia-García},
  title         = {Zero-shot Reading Comprehension and Reasoning for Spanish with {BERTIN} {GPT-J-6B}},
  date          = {2022-09},
  booktitle     = {Proceedings of the Iberian Languages Evaluation Forum (IberLEF 2022)},
  booktitleaddon = {Co-located with the Conference of the Spanish Society for Natural Language Processing (SEPLN 2022)},
  eventdate     = {2022-09-20/2022-09-25},
  venue         = {A Coru\~{n}a, Spain},
  publisher     = {CEUR Workshop Proceedings},
  year = {2022},
}

@article{hu2021lora,
  title={Lora: Low-rank adaptation of large language models},
  author={Hu, Edward J and Shen, Yelong and Wallis, Phillip and Allen-Zhu, Zeyuan and Li, Yuanzhi and Wang, Shean and Wang, Lu and Chen, Weizhu},
  journal={arXiv preprint arXiv:2106.09685},
  year={2021}
}

@article{dettmers2024qlora,
  title={Qlora: Efficient finetuning of quantized llms},
  author={Dettmers, Tim and Pagnoni, Artidoro and Holtzman, Ari and Zettlemoyer, Luke},
  journal={Advances in Neural Information Processing Systems},
  volume={36},
  year={2024}
}

@article{wang2022interpretability,
  title={Interpretability in the wild: a circuit for indirect object identification in gpt-2 small},
  author={Wang, Kevin and Variengien, Alexandre and Conmy, Arthur and Shlegeris, Buck and Steinhardt, Jacob},
  journal={arXiv preprint arXiv:2211.00593},
  year={2022}
}

@article{conmy2023towards,
  title={Towards automated circuit discovery for mechanistic interpretability},
  author={Conmy, Arthur and Mavor-Parker, Augustine and Lynch, Aengus and Heimersheim, Stefan and Garriga-Alonso, Adri{\`a}},
  journal={Advances in Neural Information Processing Systems},
  volume={36},
  pages={16318--16352},
  year={2023}
}

@article{olsson2022context,
  title={In-context learning and induction heads},
  author={Olsson, Catherine and Elhage, Nelson and Nanda, Neel and Joseph, Nicholas and DasSarma, Nova and Henighan, Tom and Mann, Ben and Askell, Amanda and Bai, Yuntao and Chen, Anna and others},
  journal={arXiv preprint arXiv:2209.11895},
  year={2022}
}

@article{meng2023memit,
  title={Mass Editing Memory in a Transformer},
  author={Kevin Meng and Sen Sharma, Arnab and Alex Andonian and Yonatan Belinkov and David Bau},
  journal={The Eleventh International Conference on Learning Representations (ICLR)},
  year={2023}
}

@inproceedings{Da2021AnalyzingCE,
  title={Analyzing Commonsense Emergence in Few-shot Knowledge Models},
  author={Jeff Da and Ronan Le Bras and Ximing Lu and Yejin Choi and Antoine Bosselut},
  booktitle={Conference on Automated Knowledge Base Construction},
  year={2021},
  url={https://api.semanticscholar.org/CorpusID:235657379}
}

@article{elhage2021mathematical,
  title={A mathematical framework for transformer circuits},
  author={Elhage, Nelson and Nanda, Neel and Olsson, Catherine and Henighan, Tom and Joseph, Nicholas and Mann, Ben and Askell, Amanda and Bai, Yuntao and Chen, Anna and Conerly, Tom and others},
  journal={Transformer Circuits Thread},
  volume={1},
  pages={1},
  year={2021}
}

@article{li2022unified,
  title={A unified understanding of deep nlp models for text classification},
  author={Li, Zhen and Wang, Xiting and Yang, Weikai and Wu, Jing and Zhang, Zhengyan and Liu, Zhiyuan and Sun, Maosong and Zhang, Hui and Liu, Shixia},
  journal={IEEE Transactions on Visualization and Computer Graphics},
  volume={28},
  number={12},
  pages={4980--4994},
  year={2022},
  publisher={IEEE}
}

@book{brown2006encyclopedia,
  title={Encyclopedia of language and linguistics (2nd Edition)},
  author={Brown, Keith},
  volume={1},
  year={2006},
  publisher={Elsevier}
}

@book{hirst1987semantic,
  title={Semantic interpretation and the resolution of ambiguity},
  author={Hirst, Graeme},
  year={1987},
  publisher={Cambridge University Press}
}

@article{partee1984compositionality,
  title={Compositionality},
  author={Partee, Barbara and others},
  journal={Varieties of formal semantics},
  volume={3},
  pages={281--311},
  year={1984},
  publisher={Foris Dordrecht}
}

@article{heimersheim2024use,
  title={How to use and interpret activation patching},
  author={Heimersheim, Stefan and Nanda, Neel},
  journal={arXiv preprint arXiv:2404.15255},
  year={2024}
}

@inproceedings{chan2022causal,
  title={Causal scrubbing: A method for rigorously testing interpretability hypotheses},
  author={Chan, Lawrence and Garriga-Alonso, Adria and Goldowsky-Dill, Nicholas and Greenblatt, Ryan and Nitishinskaya, Jenny and Radhakrishnan, Ansh and Shlegeris, Buck and Thomas, Nate},
  booktitle={AI Alignment Forum},
  pages={1828--1843},
  year={2022}
}

@article{nanda2023attribution,
  title={Attribution patching: Activation patching at industrial scale},
  author={Nanda, Neel},
  journal={URL: https://www. neelnanda. io/mechanistic-interpretability/attribution-patching},
  year={2023}
}

@inproceedings{li2024pmet,
  title={Pmet: Precise model editing in a transformer},
  author={Li, Xiaopeng and Li, Shasha and Song, Shezheng and Yang, Jing and Ma, Jun and Yu, Jie},
  booktitle={Proceedings of the AAAI Conference on Artificial Intelligence},
  volume={38},
  number={17},
  pages={18564--18572},
  year={2024}
}

@article{BERTIN,
     author={De la Rosa, Javier and Ponferrada, Eduardo G and Villegas, Paulo and Salas, Pablo Gonzalez de Prado and Romero, Manu and Grandury, Mar{\i}a},
    title = {{BERTIN}: Efficient Pre-Training of a Spanish Language Model using Perplexity Sampling},
    journal = {Procesamiento del Lenguaje Natural},
    volume = {68},
    number = {0},
    year = {2022},
    issn = {1989-7553},
    url = {http://journal.sepln.org/sepln/ojs/ojs/index.php/pln/article/view/6403},
    pages = {13--23}
}

@misc{mesh-transformer-jax,
  author = {Wang, Ben},
  title = {{Mesh-Transformer-JAX: Model-Parallel Implementation of Transformer Language Model with JAX}},
  howpublished = {\url{https://github.com/kingoflolz/mesh-transformer-jax}},
  year = 2021,
  month = May
}

@article{pile,
  title={The {P}ile: An 800GB Dataset of Diverse Text for Language Modeling},
  author={Gao, Leo and Biderman, Stella and Black, Sid and Golding, Laurence and Hoppe, Travis and Foster, Charles and Phang, Jason and He, Horace and Thite, Anish and Nabeshima, Noa and Presser, Shawn and Leahy, Connor},
  journal={arXiv preprint arXiv:2101.00027},
  year={2020}
}

@inproceedings{silva-etal-2016-categorization,
    title = "Categorization of Semantic Roles for Dictionary Definitions",
    author = "Silva, Vivian  and
      Handschuh, Siegfried  and
      Freitas, Andr{\'e}",
    editor = "Zock, Michael  and
      Lenci, Alessandro  and
      Evert, Stefan",
    booktitle = "Proceedings of the 5th Workshop on Cognitive Aspects of the Lexicon ({C}og{AL}ex - V)",
    month = dec,
    year = "2016",
    address = "Osaka, Japan",
    publisher = "The COLING 2016 Organizing Committee",
    url = "https://aclanthology.org/W16-5323",
    pages = "176--184",
}

@inproceedings{geva-etal-2021-transformer,
    title = "Transformer Feed-Forward Layers Are Key-Value Memories",
    author = "Geva, Mor  and
      Schuster, Roei  and
      Berant, Jonathan  and
      Levy, Omer",
    editor = "Moens, Marie-Francine  and
      Huang, Xuanjing  and
      Specia, Lucia  and
      Yih, Scott Wen-tau",
    booktitle = "Proceedings of the 2021 Conference on Empirical Methods in Natural Language Processing",
    month = nov,
    year = "2021",
    address = "Online and Punta Cana, Dominican Republic",
    publisher = "Association for Computational Linguistics",
    url = "https://aclanthology.org/2021.emnlp-main.446",
    doi = "10.18653/v1/2021.emnlp-main.446",
    pages = "5484--5495",
}

@inproceedings{dai-etal-2022-knowledge,
    title = "Knowledge Neurons in Pretrained Transformers",
    author = "Dai, Damai  and
      Dong, Li  and
      Hao, Yaru  and
      Sui, Zhifang  and
      Chang, Baobao  and
      Wei, Furu",
    editor = "Muresan, Smaranda  and
      Nakov, Preslav  and
      Villavicencio, Aline",
    booktitle = "Proceedings of the 60th Annual Meeting of the Association for Computational Linguistics (Volume 1: Long Papers)",
    month = may,
    year = "2022",
    address = "Dublin, Ireland",
    publisher = "Association for Computational Linguistics",
    url = "https://aclanthology.org/2022.acl-long.581",
    doi = "10.18653/v1/2022.acl-long.581",
    pages = "8493--8502",
}

@inproceedings{miller-1994-wordnet,
    title = "{W}ord{N}et: A Lexical Database for {E}nglish",
    author = "Miller, George A.",
    booktitle = "{H}uman {L}anguage {T}echnology: Proceedings of a Workshop held at {P}lainsboro, {N}ew {J}ersey, {M}arch 8-11, 1994",
    year = "1994",
    url = "https://aclanthology.org/H94-1111",
}

@inproceedings{geva-etal-2022-transformer,
    title = "Transformer Feed-Forward Layers Build Predictions by Promoting Concepts in the Vocabulary Space",
    author = "Geva, Mor  and
      Caciularu, Avi  and
      Wang, Kevin  and
      Goldberg, Yoav",
    editor = "Goldberg, Yoav  and
      Kozareva, Zornitsa  and
      Zhang, Yue",
    booktitle = "Proceedings of the 2022 Conference on Empirical Methods in Natural Language Processing",
    month = dec,
    year = "2022",
    address = "Abu Dhabi, United Arab Emirates",
    publisher = "Association for Computational Linguistics",
    url = "https://aclanthology.org/2022.emnlp-main.3",
    doi = "10.18653/v1/2022.emnlp-main.3",
    pages = "30--45",
}

@inproceedings{ferrando-etal-2023-explaining,
    title = "Explaining How Transformers Use Context to Build Predictions",
    author = "Ferrando, Javier  and
      G{\'a}llego, Gerard I.  and
      Tsiamas, Ioannis  and
      Costa-juss{\`a}, Marta R.",
    editor = "Rogers, Anna  and
      Boyd-Graber, Jordan  and
      Okazaki, Naoaki",
    booktitle = "Proceedings of the 61st Annual Meeting of the Association for Computational Linguistics (Volume 1: Long Papers)",
    month = jul,
    year = "2023",
    address = "Toronto, Canada",
    publisher = "Association for Computational Linguistics",
    url = "https://aclanthology.org/2023.acl-long.301",
    doi = "10.18653/v1/2023.acl-long.301",
    pages = "5486--5513",
}

@misc{openai2023chatgpt,
  author    = {{OpenAI}},
  title     = {{ChatGPT}: Optimizing Language Models for Dialogue},
  year      = {2023},
  url       = {https://openai.com/blog/chatgpt},
}

@inproceedings{tenney2019bert,
  title     = {{BERT} Rediscovers the Classical {NLP} Pipeline},
  author    = {Tenney, Ian and Das, Dipanjan and Pavlick, Ellie},
  booktitle = {Proceedings of the 57th Annual Meeting of the Association for Computational Linguistics},
  pages     = {4593--4601},
  year      = {2019},
  publisher = {Association for Computational Linguistics}
}

@inproceedings{vig2020investigating,
  title     = {Investigating Gender Bias in Language Models Using Causal Mediation Analysis},
  author    = {Vig, Jesse and Gehrmann, Sebastian and Belinkov, Yonatan and Qian, Sharon and Nevo, Daniel and Singer, Yaron and Shieber, Stuart},
  booktitle = {Advances in Neural Information Processing Systems},
  volume    = {33},
  pages     = {12388--12401},
  year      = {2020}
}

@inproceedings{noraset2017definition,
  title     = {Definition Modeling: Learning to Define Word Embeddings in Natural Language},
  author    = {Noraset, Thanapon and Liang, Chen and Birnbaum, Larry and Downey, Doug},
  booktitle = {Proceedings of the AAAI Conference on Artificial Intelligence},
  volume    = {31},
  pages     = {3259--3266},
  year      = {2017}
}

@inproceedings{bevilacqua2020generationary,
  title     = {Generationary or ``{H}ow We Went beyond Word Sense Inventories and Learned to Gloss''},
  author    = {Bevilacqua, Michele and Maru, Marco and Navigli, Roberto},
  booktitle = {Proceedings of the 2020 Conference on Empirical Methods in Natural Language Processing},
  pages     = {7207--7221},
  year      = {2020},
  publisher = {Association for Computational Linguistics}
}

@inproceedings{ferrando-voita-2024-information,
    title = "Information Flow Routes: Automatically Interpreting Language Models at Scale",
    author = "Ferrando, Javier  and
      Voita, Elena",
    editor = "Al-Onaizan, Yaser  and
      Bansal, Mohit  and
      Chen, Yun-Nung",
    booktitle = "Proceedings of the 2024 Conference on Empirical Methods in Natural Language Processing",
    month = nov,
    year = "2024",
    address = "Miami, Florida, USA",
    publisher = "Association for Computational Linguistics",
    url = "https://aclanthology.org/2024.emnlp-main.965/",
    doi = "10.18653/v1/2024.emnlp-main.965",
    pages = "17432--17445"
}

@article{xu-etal-2024-tip,
  title={On the tip of the tongue: Analyzing conceptual representation in large language models with reverse-dictionary probe},
  author={Xu, Ningyu and Zhang, Qi and Zhang, Menghan and Qian, Peng and Huang, Xuanjing},
  journal={arXiv preprint arXiv:2402.14404},
  year={2024}
}

@article{geiger2021causal,
  title={Causal abstractions of neural networks},
  author={Geiger, Atticus and Lu, Hanson and Icard, Thomas and Potts, Christopher},
  journal={Advances in neural information processing systems},
  volume={34},
  pages={9574--9586},
  year={2021}
}
